\documentclass[10pt,twocolumn,letterpaper]{article}

\usepackage{bm}
\usepackage{iccv}
\usepackage{times}
\usepackage{epsfig}
\usepackage{graphicx}
\usepackage{amsmath}
\usepackage{amssymb}
\usepackage{subfigure}
\usepackage{amsthm}
\usepackage{mathrsfs}
\usepackage{algorithmic}
\usepackage{verbatim}
\usepackage{booktabs}
\usepackage{colortbl}
\usepackage{caption}
\usepackage{authblk}
\usepackage{url}
\usepackage{multirow}
\usepackage[title]{appendix}

\usepackage[linesnumbered,boxed]{algorithm2e}

\def\mycolor{\cellcolor[rgb]{0.95,0.95,0.95}}

\usepackage[pagebackref=true,breaklinks=true,letterpaper=true,colorlinks,bookmarks=false]{hyperref}

\iccvfinalcopy 

\def\iccvPaperID{****} 

\begin{document}

\title{Location-Sensitive Visual Recognition with Cross-IOU Loss}


\author{Kaiwen Duan$^{1}$ ~~ Lingxi Xie$^{2}$ ~~ Honggang Qi$^{1}$ ~~
Song Bai$^{3}$ ~~ Qingming Huang$^{1}$ ~~ Qi Tian$^{2}$ \\
$^{1}$University of Chinese Academy of Sciences ~~ $^{2}$Huawei Inc \\
$^{3}$Huazhong University of Science and Technology\\
{\tt\small kaiwenduan@outlook.com, \tt\small 198808xc@gmail.com, \tt\small \{hgqi,qmhuang\}@ucas.ac.cn \\
\tt\small songbai.site@gmail.com, \tt\small tian.qi1@huawei.com}
}

\maketitle
\ificcvfinal\thispagestyle{empty}\fi

\begin{abstract}
Object detection, instance segmentation, and pose estimation are popular visual recognition tasks which require localizing the object by internal or boundary landmarks. This paper summarizes these tasks as \textbf{location-sensitive} visual recognition and proposes a unified solution named location-sensitive network (\textbf{LSNet}). Based on a deep neural network as the backbone, LSNet predicts an anchor point and a set of landmarks which together define the shape of the target object. The key to optimizing the LSNet lies in the ability of fitting various scales, for which we design a novel loss function named \textbf{cross-IOU loss} that computes the cross-IOU of each anchor-landmark pair to approximate the global IOU between the prediction and ground-truth. The flexibly located and accurately predicted landmarks also enable LSNet to incorporate richer contextual information for visual recognition. Evaluated on the MS-COCO dataset, LSNet set the new state-of-the-art accuracy for anchor-free object detection (a $53.5\%$ box AP) and instance segmentation (a $40.2\%$ mask AP), and shows promising performance in detecting multi-scale human poses. Code is available at \url{https://github.com/Duankaiwen/LSNet}.
\end{abstract}


\section{Introduction}

Object recognition is a fundamental task in computer vision. Beyond image classification~\cite{deng2009imagenet} that depicts an image using a single semantic label, there exist other recognition tasks that not only predict the class of the object but also localize it using fine-scaled information. In this paper, we consider three popular examples including object detection~\cite{everingham2010pascal,lin2014microsoft}, instance segmentation~\cite{everingham2010pascal,cordts2016cityscapes}, and human pose estimation~\cite{andriluka20142d,lin2014microsoft}. We notice that, despite the fact that the rapid progress of deep learning~\cite{lecun2015deep} has introduced powerful deep networks as the backbone~\cite{he2016deep,xie2017aggregated,gao2019res2net}, the designs of the \textit{head} module for detection~\cite{liu2016ssd,redmon2017yolo9000,ren2015faster,tian2019fcos,zhang2019freeanchor,law2019cornernet,law2018cornernet,yu2016unitbox,carion2020end}, segmentation~\cite{xie2020polarmask,he2017mask,peng2020deep,chen2019hybrid}, and pose estimation~\cite{sun2019deep,wang2020deep,cao2019openpose,newell2016stacked,zhou2019objects} have fallen into individual sub-fields. This is mainly due to the difference in the prediction target, \textit{i.e.}, a \textit{bounding box} for detection, a pixel-level \textit{mask} for segmentation, and a set of \textit{keypoints} for pose estimation, respectively.

\begin{figure}[!t]
	\centering 
	\includegraphics[width = 8.3cm]{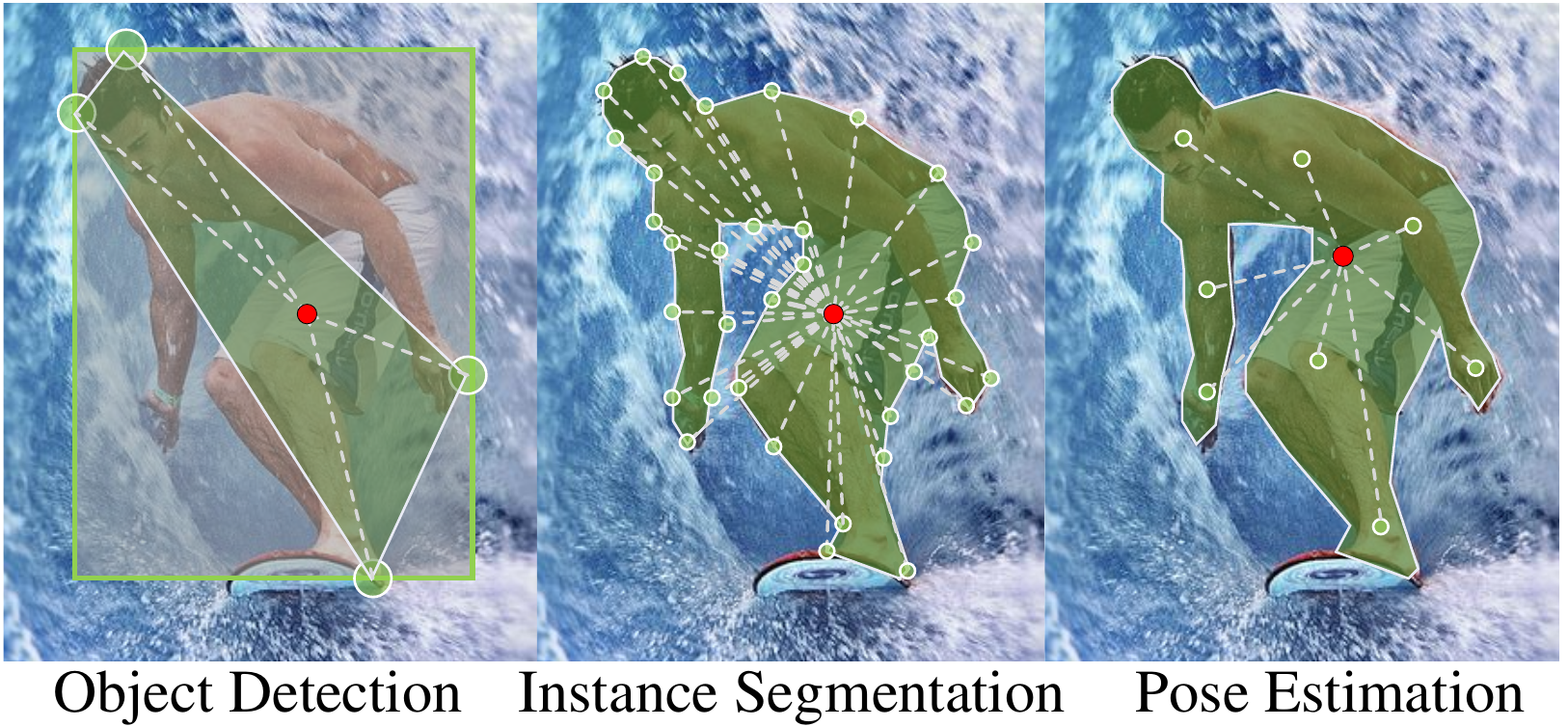}
	\caption{The location-sensitive visual recognition tasks, including object detection, instance segmentation, and human pose estimation, can be formulated into localizing an anchor point (in \textcolor{red}{red}) and a set of landmarks (in \textcolor{green}{green}). Our work aims to offer a unified framework for these tasks.}
	\label{fig:introduction}
	\vspace{1ex}
\end{figure}

Going one step further, we merge the aforementioned three tasks into one, named the \textbf{location-sensitive visual recognition (LSVR)}. On the basis of the definition, we propose a location-sensitive network (\textbf{LSNet}) as a unified formulation to deal with them all. The LSNet is built upon any network backbone, \textit{e.g.}, those designed for image classification. The key is to relate an object to an anchor point and a set of landmarks that accurately localize the object. In particular, the landmarks should correspond to the four extreme points for object detection, sufficiently dense boundary pixels for instance segmentation, and the keypoints for human pose estimation. Note that the anchor point as well as landmarks can be also used for extracting discriminative features of the object and thus assisting recognition. Figure~\ref{fig:introduction} illustrates the overall idea.

The major difficulty of optimizing the LSNet lies in the requirement of fitting objects of different scales and different properties, which existing methods including the smooth-$\ell_1$ loss and the IOU loss suffer cannot satisfy. This motivates us to design a novel loss function named the \textbf{cross-IOU loss}. It assumes that the landmarks are uniformly distributed around the anchor point and thus approximates the IOU (between the prediction and ground-truth) using the coordinate of the offset vectors. The cross-IOU loss is easily implemented in a few lines of codes. Compared to other loss functions, it achieves a better trade-off between the global and local properties and transplants easier to multi-scale feature maps without specific parameter tuning.

We perform all three tasks on the MS-COCO dataset~\cite{lin2014microsoft}. LSNet, equipped with the cross-IOU loss, achieves competitive recognition accuracy. We further equip the LSNet with a pyramid of deformable convolution that extracts discriminative visual cues around the landmarks. As a result, LSNet reports a $53.5\%$ box AP and a $40.2\%$ mask AP, both of which surpass all existing anchor-free methods. For human pose estimation, LSNet reports competitive results without using the heatmaps, offering a new possibility to the community. Moreover, LSNet shows a promising ability in detecting human poses in various scales, some of which were not annotated in the dataset.

On top of these results, we claim two-fold contributions of this work. \textbf{First}, we present the formulation of location-sensitive visual recognition that inspires the community to consider the common property of these tasks. \textbf{Second}, we propose the LSNet as a unified framework in which the key technical contribution is the cross-IOU loss.

\vspace{1ex}
\section{Related Work}
\textbf{Deep neural networks} have been widely applied for visual recognition tasks. Among them, \textbf{image classification}~\cite{deng2009imagenet} is the fundamental task that facilitates the design of powerful network backbones~\cite{he2016deep,xie2017aggregated,gao2019res2net}. Beyond image-level description, there exist fine-scaled tasks, including object detection, instance segmentation, and pose estimation, which focus on depicting different aspects of the object. For example, the bounding boxes locate objects simply and efficiently but lack the details, while masks and keypoints reflect the shape and pose of the objects but usually need the bounding boxes to locate object firstly. According to the different properties of different tasks, many representative methods have been developed.

The \textbf{object detection} methods can be roughly categorized into anchor-based and anchor-free. The anchor-based methods detect objects by placing a pre-defined set of anchor boxes, predicting the class and score for each anchor, and finally regressing the preserved boxes tightly around the objects. The representative methods include Fast R-CNN~\cite{girshick2015fast}, Faster RCNN~\cite{ren2015faster}, R-FCN~\cite{dai2016r}, SSD~\cite{liu2016ssd}, RetinaNet~\cite{lin2017focal}, Cascade R-CNN~\cite{cai2018cascade}, \textit{etc}. On another line, the anchor-free methods usually represent an object as a combination of geometry. Among them, CornerNet~\cite{law2018cornernet} and DeNet~\cite{tychsen2017denet} generated a bounding box by predicting a pair of corner keypoints, and CenterNet (keypoint triplets)~\cite{duan2019centernet} and CPNDet~\cite{duan2020corner} applied semantic information within the objects to filter out the incorrect corner pairs. FCOS~\cite{tian2019fcos}, RepPoints~\cite{yang2019reppoints}, FoveaBox~\cite{kong2020foveabox}, SAPD~\cite{zhu2019soft}, CenterNet (objects as points)~\cite{zhou2019objects}, YOLO~\cite{redmon2016you}, \textit{etc.}, defined a bounding box by placing a single point (called anchor point) within the object and predicting its distances to the object boundary.

For \textbf{instance segmentation}, there are mainly two kinds of methods, namely, the pixel-based and the contour-based methods. The pixel-based methods consider the segmentation problem as predicting the class of each single pixel. One of the representative work is Mask R-CNN~\cite{he2017mask}, which first predicted the bounding box to help locating the objects, and then used pixel-wise classification to determine the object mask. The contour-based methods instead represent an object by the contour. They often start with a set points that are roughly located around the object boundary, and the points gradually get closer to the object boundary under iteration. The early representative methods are the snake series~\cite{kass1988snakes,cohen1991active,gunn1997robust,cootes1995active} and the recent efforts include using deep neural network~\cite{girshick2014rich} and the idea of anchor-free to improve the features, such as DeepSnake~\cite{peng2020deep} and PolarMask~\cite{xie2020polarmask}.

There are two mainstreams for \textbf{human pose estimation}, namely, the bottom-up methods~\cite{cai2020learning,cheng2020higherhrnet,insafutdinov2016deepercut,newell2017associative,cao2019openpose} and top-down methods~\cite{zhou2019objects,toshev2014deeppose,wei2016convolutional,liu2020improving}. The bottom-up methods first detect the human parts and then locates the keypoints in each object, while the top-down first locates all the keypoints in the human body and then composes the individual parts into a person. The keypoints are often sparsely distributed in an image and thus are difficult to be accurately located. A practical solution is to detect the keypoints in a high-resolution feature map, called heatmap~\cite{newell2016stacked,cheng2020higherhrnet,cheng2020higherhrnet}. However, applying the heatmap makes the  optimization hard and introduces a complex post-processing operation. CenterNet~\cite{zhou2019objects} proposes a neat and simple method, which only predicts a center heatmap and the keypoints are obtained by regressing the vector from the center within objects to the keypoints.


This paper particularly focuses on the \textbf{anchor-free methods} for visual recognition. These methods originated from object detection and have drawn a lot of attention recently. They do not rely on the pre-defined anchor boxes to locate objects but by points and distance. 
Therefor, the anchor-free methods enjoy the ability to extend into any directions. This offers the researchers a possibility to unify the visual recognition tasks. Recent trends have spent effort in extending the anchor-free methods in object detection into other tasks, \textit{e.g.}, PolarMask~\cite{xie2020polarmask} tries to extend the anchor-free into instance segmentation, while CenterNet~\cite{zhou2019objects} applies it into the pose estimation. Compared with our framework, both of them have limitations, which we will give a detailed discussion in section~\ref{approach:framework}.

\begin{figure*}[!t]
\centering 
\includegraphics[width=1\textwidth]{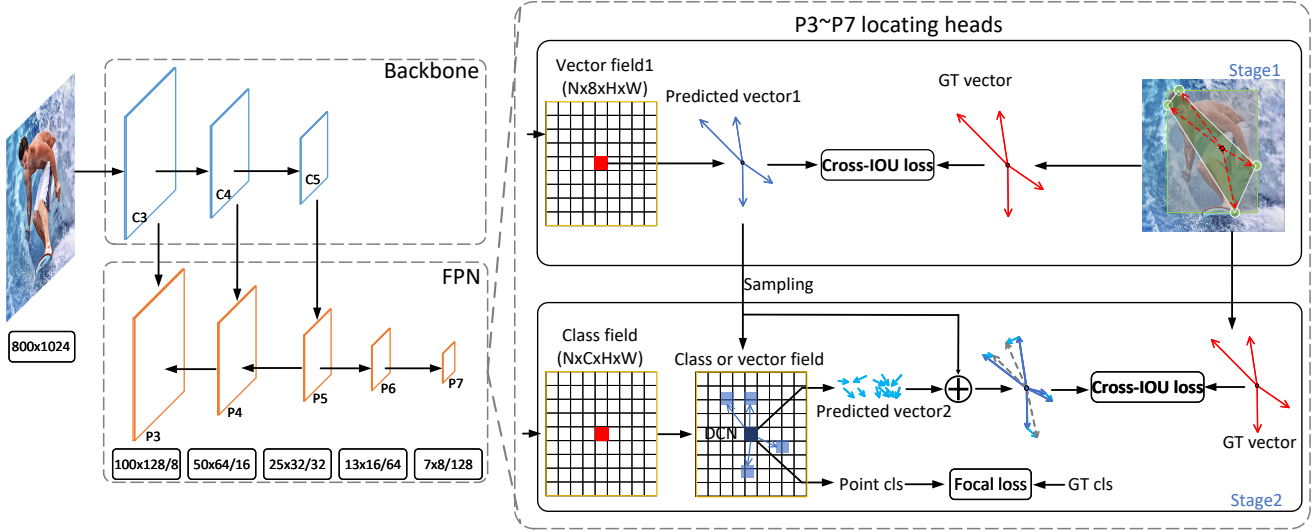}
\caption{An illustration of the overall framework of LSNet, using object detection as an example, \textit{i.e.}, $N=4$. In the left part, C3--C5 denote the feature maps in the backbone that are connected to the feature pyramid~\cite{lin2017feature}, and P3--P7 denote the FPN layers used for final prediction. LSNet is partitioned into two stages. In the first stage, we predict a set of offset vectors that relate the anchor point to the extreme points under the supervision of the cross-IOU loss. In the second stage, the predicted vectors are used as the offsets of the deformable convolution (DCN)~\cite{dai2017deformable} to extract complementary visual features around the extreme points. These features are used for refining the localization and predicting the object class.}
\label{fig:architecture}
\end{figure*}

\section{Our Approach}
\label{approach}

\subsection{Location-Sensitive Visual Recognition}
\label{approach:problem}

Visual recognition tasks start with an image, $\mathbf{X}$. Image classification aims to assign a class label $c$ for the entire image, yet there are more challenging tasks for fine-scaled recognition. These tasks often focus on the instances (\textit{i.e.}, individual objects) in the image and depict the object properties from different aspects. Typical examples include object detection that uses a rectangular box that tightly cover the object, instance segmentation that finds out each pixel that belongs to the object, and human pose estimation that localizes the landmarks of the object (\textit{i.e.}, human keypoints). We use $\mathbf{b}\in\mathbb{R}^4$, $\mathbf{s}\in[0,1]^{W\times H}$, and $\mathbf{k}\in\mathbb{R}^{K\times2}$ to indicate the target of these tasks, where $W$ and $H$ are the image width and height, and $K$ is the number of keypoints.

An important motivation of our work is that, although these tasks differ from each other in the form of description, they share the common requirement that the model should be sensitive to the location of the anchor and/or landmarks. Throughout the remaining part of this paper, we refer to these tasks as \textbf{location-sensitive} visual recognition and design a unified framework for them.

\subsection{LSNet: A Unified Framework}
\label{approach:framework}

The proposed location-sensitive network (\textbf{LSNet}) starts with a backbone (\textit{e.g.}, the ResNet~\cite{he2016deep}, ResNeXt~\cite{xie2017aggregated}, \textit{etc.}) that extracts features from the input image. We denote the process using $\tilde{\mathbf{x}}=f(\mathbf{X})$. Next, an anchor point and a few landmarks are predicted on top of $\tilde{\mathbf{x}}$, denoted as $\mathbf{p}=g(\tilde{\mathbf{x}})$. Here we define $\mathbf{p}\in\mathbb{R}^{(N+1)\times2}$ where $N$ is the number of landmarks, $\mathbf{p}_0\in\mathbb{R}^2$ is the anchor point, and $\mathbf{p}_n\in\mathbb{R}^2$ is a landmark for $n=1,2,\ldots,N$.

As a unified framework, the key is to relate the prediction targets (\textit{i.e.}, $\mathbf{b}\in\mathbb{R}^4$, $\mathbf{s}\in[0,1]^{W\times H}$, and $\mathbf{k}\in\mathbb{R}^{K\times2}$ as aforementioned) to $\mathbf{p}$. For \textbf{object detection}, this is done by finding an extreme point (a pixel that belongs to the object and is tangent to the bounding box) on each edge of the bounding box\footnote{As a disclaimer, there may exist multiple or even continuous extreme points on each edge. We assume the method to find any one of them, by which it confirms the prediction of the bounding box.}, \textit{i.e.}, $N=4$. For \textbf{instance segmentation}, we locate a fixed number (\textit{e.g.}, $36$ in the experiments, $N=36$) of landmarks along the contour and thus use the formed polygon to approximate the shape of the object\footnote{In case that the mask is not \textit{simply-connected} in topology, we follow PolarMask~\cite{xie2020polarmask} to deal with each part separately and ignore the holes.}. For \textbf{human pose estimation}, we follow the definition of the dataset to learn a fixed number of keypoints, \textit{e.g.}, in the MS-COCO dataset, $N=17$.

Figure~\ref{fig:architecture} shows the pipeline of LSNet. It belongs to the category of anchor-free methods, \textit{i.e.}, there is no need to pre-define a set of anchor boxes for localizing the object. LSNet is partitioned into two stages, where the first stage predicts an anchor point from the FPN head and relates it with a set of landmarks, and the second stage composes the landmarks into an object with the desired geometry (\textit{e.g.}, a bounding box). To facilitate accurate localization, we use the ATSS assigner~\cite{zhang2020bridging} to assign more anchor points for each object and extract features with deformable convolution (DCN) upon the predicted landmarks. The entire model is an end-to-end trainable learnable function.


LSNet receives two sources of supervision for localization and classification, elaborated in Sections~\ref{approach:localization} and~\ref{approach:classification}, respectively. The localization loss is added to both stages, where the major contribution is a unified loss that fits the properties of different tasks, and the classification loss is added to the second stage upon the DCN features.

LSNet extends the border of anchor-free methods for location-sensitive visual recognition. We briefly review two counterparts. \textbf{(i)} CenterNet predicted horizontal or vertical offsets beyond the anchor point for object detection. This limits its ability in finding the extreme points and extracting discriminative features, yet it cannot perform instance segmentation. \textbf{(ii)} PolarMask used a polar coordinate system for instance segmentation, making it difficult to process the situation that a ray intersects the object multiple times at some direction. In comparison, LSNet easily handles the challenging scenarios and report superior performance (see the experimental part, section~\ref{Experiments}).


\subsection{Cross IOU Loss}
\label{approach:localization}

The unified framework raises new challenges to the supervision of localization, because the function needs to consider both the global and local properties of the object. To clarify, we notice that the evaluation of object detection and instance segmentation judges if an object is correctly recognized by \textit{the global IOU} between the prediction and the ground-truth, while pose estimation measures the accuracy by each \textit{individual keypoint}.

To this end, we design the \textbf{cross-IOU loss} as the unified supervision. The loss is defined upon the predicted and ground-truth objects, and we use $\mathbf{p}^\star$, where $\mathbf{p}^\star=(\mathbf{p}^\star_x, \mathbf{p}^\star_y)$, to denote the corresponding ground-truth of the anchor point and landmarks. We compute the offset from the anchor point to the landmarks in a \textbf{cross-coordinate system}, \textit{i.e.}, $\mathbf{q}_n=\left[\left(p_{n,x}\!-\!p_{0,x}\right)^-\!,\left(p_{n,x}\!-\!p_{0,x}\right)^+\!,\left(p_{n,y}\!-\!p_{0,y}\right)^-\!,\left(p_{n,y}\!-\!p_{0,y}\right)^+\!\right]$ for $n=1,2,\ldots,N$, where $\left(a\right)^-$ and $\left(a\right)^+$ denotes $\max\{-a,0\}$ and $\max\{a,0\}$, respectively. Finally, we write the cross-IOU loss as:
\begin{equation}
\label{eqn:cIOU}
\mathrm{cIOU}\!\left(\mathbf{q}_n,\mathbf{q}_n^\star\right)=\frac{\left|\min\left\{\mathbf{q}_n-\mathbf{q}_n^\star\right\}\right|_1}{\left|\max\left\{\mathbf{q}_n-\mathbf{q}_n^\star\right\}\right|_1},
\end{equation}
where $\left|\cdot\right|_1$ indicates the $\ell_1$-norm. In other words, the cross-IOU function rewards the components ($\mathbf{q}_n$ and $\mathbf{q}_n^\star$) of similar length (in which case the prediction and the ground-truth maximally overlap) and penalizes the components on different directions. Based on the Eqn~\eqref{eqn:cIOU}, we define the cross-IOU loss as $\mathcal{L}_\mathrm{cIOU}=1-\frac{1}{n}\sum_{n=1}^N\mathrm{cIOU}\!\left(\mathbf{q}_n,\mathbf{q}_n^\star\right)$. Obviously, when $\mathbf{p}_n=\mathbf{p}_n^\star$ for all $n$, we have $\mathcal{L}_\mathrm{cIOU}=0$ as expected\footnote{The current form of $\mathcal{L}_\mathrm{cIOU}$ can cause the gradients over $p_{n,x}$ and $p_{n,y}$ to be $0$ when the corresponding dimensions of prediction and ground-truth are of different signs. We design a softened prediction mechanism to solve the issue. Please refer to the Appendix~\ref{appendix_a} for details.}.

The cross-IOU loss brings a direct benefit that it fits different scales of features without the need of specific parameter tuning. This alleviates the difficulty of integrating multi-scale information, \textit{e.g.}, using the feature pyramid~\cite{lin2017feature}. In comparison, the smooth-$\ell_1$ loss~\cite{girshick2015fast} is sensitive to the scale of vector (\textit{e.g.}, the loss value tends to be large when the feature resolution is large) and neglects the relationship between the components that are from the same vector. Moreover, by approximating the IOU using individual components, the cross-IOU loss is flexibly transplanted to instance segmentation and human pose estimation, unlike the original IOU loss~\cite{yu2016unitbox} that is difficult to compute on polygons (for segmentation) and undefined for discrete keypoints (for pose estimation).

\begin{figure}[!tb]
\centering 
\includegraphics[width = 0.45\textwidth]{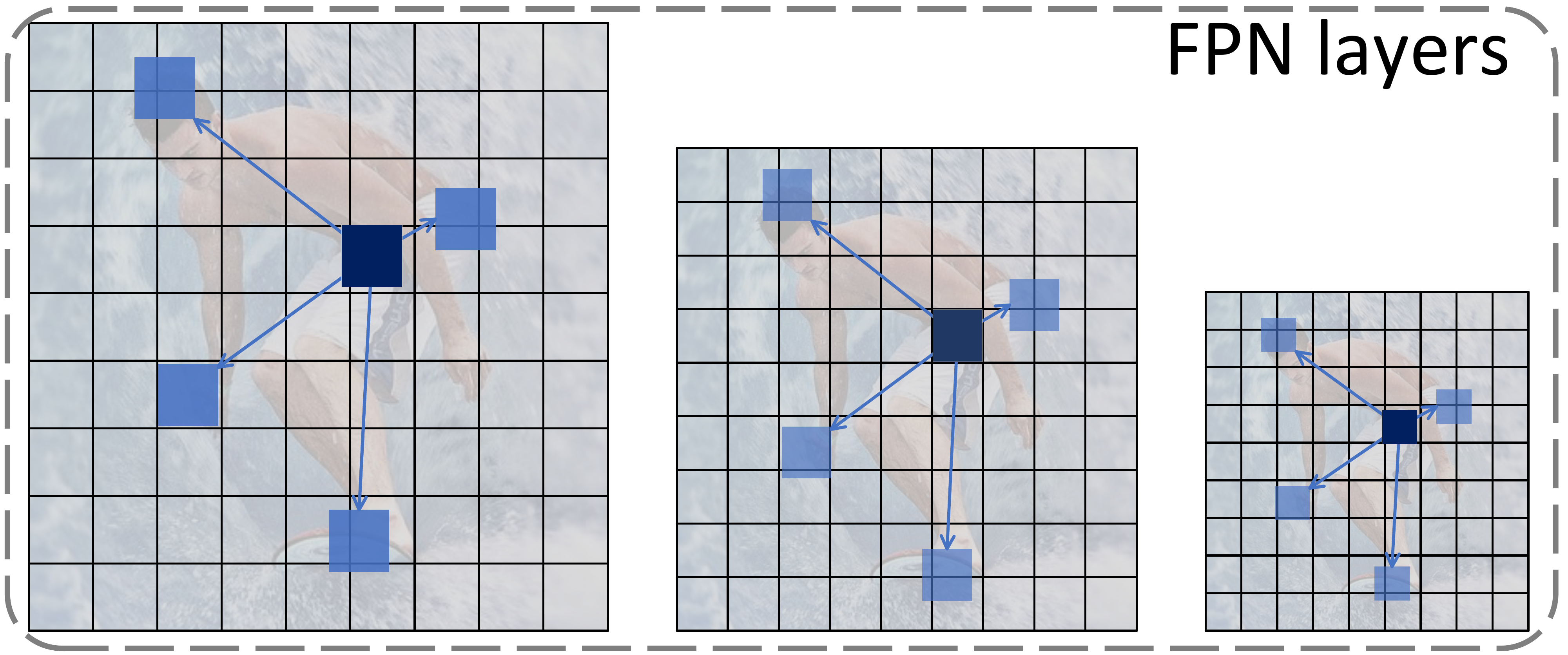}
\caption{An illustration of the Pyramid DCN structure. The DCN kernel is allowed to be rescaled to the adjacent FPN layers to extract richer features. The blue boxes and arrows denote the positions of convolution determined by the predicted offset vectors (the blue arrows).}
\label{fig:PyramidDCN}
\end{figure}

\subsection{Pyramid DCN}
\label{approach:classification}

To enhance discriminative information for recognition, we use deformable convolution (DCN)~\cite{dai2017deformable,zhu2019deformable} to extract features from the landmarks. The standard DCN has $9$ offsets, while the number of offsets is $4$, $36$, and $17$ for detection, segmentation, and pose estimation, respectively. In the latter two cases, to avoid redundant features extracted from close areas, we sample $9$ landmarks uniformly from the candidates. We further build the feature extraction module upon the feature pyramid~\cite{lin2017feature}. The offsets are adjusted to different stages by accordingly rescaling the vectors.

We name the proposed method Pyramid-DCN, and illustrate it in Figure~\ref{fig:PyramidDCN}.  As shown in experiments, both feature extraction from the landmarks and using the pyramid structure improve recognition accuracy.



\section{Experiments}
\label{Experiments}
\subsection{Dataset and Evaluation Metrics}

We evaluate our framework on the MS-COCO $2017$ dataset~\cite{lin2014microsoft}, which is a popular, large-scale object detection, segmentation, and human pose dataset. For object detection and segmentation, it contains over $118\mathrm{K}$ training images, $5\mathrm{K}$ validation images and $20\mathrm{K}$ \textit{test-dev} images covering $80$ object categories. For human pose, the person instance is labeled with $17$ keypoints, containing over $57\mathrm{K}$ training images with $150\mathrm{K}$ person instances, $5\mathrm{K}$ validation images and $20\mathrm{K}$ \textit{test-dev} images. 

The average precision (AP) metric is applied to characterize the performance of our method as well as other competitors. 
There are subtle differences in the definition of AP for different tasks. For object detection, AP is calculated the average precision under different bounding box IOU thresholds(from $0.5$ to $0.95$), while the bounding box IOU is replaced with the mask IOU in instance segmentation. In the human pose task, AP is calculated based on the object keypoint similarity (OKS), which reflects the distance between the predicted keypoints and the annotations.

\subsection{Implementation Details}

We use ResNet~\cite{he2016deep}, ResNeXt~\cite{xie2017aggregated} and Res2Net~\cite{gao2019res2net} with the weights pre-trained on ImageNet~\cite{deng2009imagenet} as our backbones, respectively. The feature pyramid network (FPN)~\cite{lin2017feature} is applied to deal with objects with different scales. For object detection, we set four vectors for each object to learn to find the four extreme points (top, left, bottom, right). We refer to ExtremeNet~\cite{zhou2019bottom} to obtain extreme point annotations from the object mask\footnote{As a side comment, when annotating the bounding boxes on a dataset, we recommend annotating an object by clicking the four extreme points (top-most, left-most, bottom-most, right-most) of the object. According to~\cite{papadopoulos2017extreme}, this way is roughly four times faster than directly annotating the bounding boxes. In addition, the extreme point itself contains the object information as well.}. For instance segmentation and human pose estimation, we set $36$ vectors for each instance to regress the location of contour points and $17$ vectors to regress the $17$ keypoints.

\noindent \textbf{Training and Inference.} We train our framework on eight NVIDIA Tesla-V100 GPUs with two images on each GPU. The initial learning rate is set as $0.01$, the weight decay as $0.0001$ and momentum as $0.9$. In the ablation study, we use a ResNet-50~\cite{he2016deep} pre-trained on ImageNet~\cite{deng2009imagenet} as the backbone, and fine-tune the model for $12$ epochs using a single-scale of $[800, 1333]$ and augment the training images with random horizontal flipping. The learning rate decays by a factor of $10$ at after the $8$th and $11$th epochs, respectively. We also use stronger backbones and longer training epochs ($24$ epochs for object detection, $30$ epochs for instance segmentation and $60$ epochs for human pose with the learning rate decayed by a factor of $10$ after the $16$th, $22$nd, and $50$th epochs, respectively) and multi-scale input images (from $[400, 1333]$ to $[960, 1333]$) to further improve the recognition accuracy. In the first stage, we only select the anchor point closest to the center of the object as a positive sample. In the second stage, we use the ATSS~\cite{zhang2020bridging} assigner to assign the anchor points for each object. The overall loss function is 
\begin{align} 
{\mathcal{L}}={\mathcal{L}_\mathrm{cls}+ \beta\cdot\mathcal{L}_\mathrm{vector_1}+ \gamma\cdot\mathcal{L}_\mathrm{vector_2}},
\label{loss} 
\end{align}
where $\mathcal{L}_\mathrm{cls}$ and $\mathcal{L}_\mathrm{vector}$ denote the Focal loss~\cite{lin2017focal} and our cross-IOU loss, respectively. We set the balancing coefficients, $\beta$ and $\gamma$, to be $1.0$ and $2.0$ in the experiments. During the inference, both the single-scale testing and multi-scale testing strategy are applied. We use the scale of $[800,1333]$ for single-scale testing. For the multi-scale testing, we refer to ATSS~\cite{zhang2020bridging} to set the image scales. We also use the non-maximum suppression (NMS) strategy with a threshold of $0.6$ to remove the redundant results.

\subsection{Object Detection}
\begin{table*}[!t]
\small
\begin{center}
\resizebox{0.9\textwidth}{!}{
\begin{tabular}{l|c|c|c|c|cccccc}
\hline
Method & Backbone & Epoch & MS$_\mathrm{train}$ & FPS & AP & AP$_{50}$ & AP$_{75}$ & AP$_\mathrm{S}$ & AP$_\mathrm{M}$ & AP$_\mathrm{L}$\\
\hline
\hline
\textbf{Anchor-based:} & & & & & & & & & &\\
Libra R-CNN~\cite{pang2019libra} & X-101-64x4d & 12 & & 8.5 & 43.0 & 64.0 & 47.0 & 25.3 & 45.6 & 54.6\\
AB+FSAF~\cite{zhu2019feature}~$\dagger$ & X-101-64x4d & 18 &$\checkmark$&-&44.6 & 65.2 & 48.6 & 29.7 & 47.1 & 54.6\\
FreeAnchor~\cite{zhang2019freeanchor}~$\dagger$&X-101-32x8d&24&$\checkmark$&-&47.3& 66.3 & 51.5 & 30.6 & 50.4 & 59.0\\
GFLV1~\cite{zhang2019freeanchor}&X-101-32x8d&24&$\checkmark$&10.7&48.2& 67.4 & 52.6 & 29.2 & 51.7 & 60.2\\
ATSS~\cite{zhang2020bridging}~$\dagger$ & X-101-64x4d-DCN &24&$\checkmark$&-&50.7 & 68.9 & 56.3 & 33.2 & 52.9 & 62.4\\
PAA~\cite{kim2020probabilistic}~$\dagger$ & X-101-64x4d-DCN &24&$\checkmark$&-&51.4&69.7 & 57.0 & 34.0 & 53.8 & 64.0\\
GFLV2~\cite{li2020generalizedv2}$\dagger$ & R2-101-DCN & 24 & $\checkmark$ & - & 53.3 & 70.9 & 59.2 & 35.7 & 56.1 & 65.6\\  
YOLOv4-P7~\cite{wang2020scaled}$\dagger$ & CSP-P7 & 450 & $\checkmark$ & - & \textbf{56.0} & \textbf{73.3} & \textbf{61.2} & \textbf{38.9} & \textbf{60.0} & \textbf{68.6}\\  
\hline
\hline
\textbf{Anchor-free:} & & & & & & & &\\
ExtremeNet~\cite{zhou2019bottom}~$\dagger$ & HG-104 & 200 &$\checkmark$&-& 43.2 & 59.8 & 46.4 & 24.1 & 46.0 & 57.1\\
RepPointsV1~\cite{yang2019reppoints}~$\dagger$ & R-101-DCN & 24&$\checkmark$&-& 46.5 & 67.4 & 50.9 & 30.3 & 49.7 & 57.1\\
SAPD~\cite{zhu2019soft} & X-101-64x4d-DCN & 24 &$\checkmark$&4.5& 47.4 & 67.4 & 51.1 & 28.1 & 50.3 & 61.5\\
CornerNet~\cite{law2018cornernet}~$\dagger$ & HG-104 & 200&$\checkmark$&-&42.1 & 57.8 & 45.3 & 20.8 & 44.8 & 56.7\\
DETR~\cite{carion2020end}& R-101 & 500 &$\checkmark$&10& 44.9 & 64.7 & 47.7 & 23.7 & 49.5 & 62.3\\
CenterNet~\cite{duan2019centernet}~$\dagger$ & HG-104 & 190 &$\checkmark$&-& 47.0 & 64.5 & 50.7 & 28.9 & 49.9 & 58.9\\
CPNDet~\cite{duan2020corner}~$\dagger$ & HG-104 & 100 &$\checkmark$&-& 49.2 & 67.4 & 53.7 & 31.0& 51.9 & 62.4\\
BorderDet~\cite{qiu2020borderdet}~$\dagger$ & X-101-64x4d-DCN&24&$\checkmark$&-& 50.3 & 68.9 & 55.2 & 32.8 & 52.8 & 62.3\\
FCOS-BiFPN~\cite{tian2020fcos}&X-101-32x8-DCN& 24&$\checkmark$&n/a& 50.4 & 68.9 & 55.0 & 33.2 & 53.0 & 62.7\\
RepPointsV2~\cite{chen2020reppoints}~$\dagger$ & X-101-64x4d-DCN & 24 &$\checkmark$&-& 52.1 & 70.1 & 57.5 & 34.5 & 54.6 & 63.6\\
\hline
\mycolor{\textbf{LSNet}} & \mycolor{R-50} & \mycolor{24} &\mycolor{$\checkmark$}& \mycolor{12.7} & \mycolor{44.8} & \mycolor{64.1} & \mycolor{48.8} & \mycolor{26.6}
 & \mycolor{47.7} & \mycolor{55.7}\\
\mycolor{\textbf{LSNet}} & \mycolor{X-101-64x4d} & \mycolor{24} &\mycolor{$\checkmark$}& \mycolor{7.2} & \mycolor{48.2} & \mycolor{67.6} & \mycolor{52.6} & \mycolor{29.6}
 & \mycolor{51.3} & \mycolor{60.5}\\
\mycolor{\textbf{LSNet}} & \mycolor{X-101-64x4d-DCN} & \mycolor{24} &\mycolor{$\checkmark$}& \mycolor{5.9} & \mycolor{49.6} & \mycolor{69.0} & \mycolor{54.1} 
& \mycolor{30.3} & \mycolor{52.8} & \mycolor{62.8}\\
\mycolor{\textbf{LSNet-CPV}} & \mycolor{X-101-64x4d-DCN} & \mycolor{24} &\mycolor{$\checkmark$}& \mycolor{5.1} & \mycolor{50.4} & \mycolor{69.4} & \mycolor{54.5} & 
\mycolor{31.0} & \mycolor{53.3} & \mycolor{64.0}\\
\mycolor{\textbf{LSNet-CPV}} & \mycolor{R2-101-DCN} & \mycolor{24} &\mycolor{$\checkmark$}& \mycolor{6.3} & \mycolor{51.1} & \mycolor{70.3} & \mycolor{55.2} & 
\mycolor{31.2} & \mycolor{54.3} & \mycolor{65.0}\\
\mycolor{\textbf{LSNet-CPV}~$\dagger$} & \mycolor{R2-101-DCN} & \mycolor{24} &\mycolor{$\checkmark$}& \mycolor{-}& \mycolor{\textbf{53.5}} & \mycolor{\textbf{71.1}} &
\mycolor{\textbf{59.2}} & \mycolor{\textbf{35.2}} & \mycolor{\textbf{56.4}} & \mycolor{\textbf{65.8}}\\
\hline
\end{tabular}}
\end{center}
\caption{A comparison between LSNet and the sate-of-the-art methods in object detection on the MS-COCO \textit{test-dev} set. LSNet surpasses all competitors in the anchor-free group. The abbreviations are: `R' -- ResNet~\cite{he2016deep}, `X' -- ResNeXt~\cite{xie2017aggregated}, `HG' -- Hourglass network~\cite{newell2016stacked}, `R2' -- Res2Net~\cite{gao2019res2net}, `CPV' -- corner point verification~\cite{chen2020reppoints}, `MS$_\mathrm{train}$' -- multi-scale training, `$\dagger$' -- multi-scale testing~\cite{zhang2020bridging}.}
\label{tab:detection}
\end{table*}
\noindent \textbf{Comparisons to SOTA.} We evaluate the detection accuracy of LSNet on the MS-COCO \textit{test-dev} set, the results are shown in Table~\ref{tab:detection}. As Table~\ref{tab:detection} shows, our method is an anchor-free detector, with a backbone of ResNet-50, LSNet achieves a box AP of $44.8 \%$ with 12.7 FPS, which has been competitive with other detectors that equipped with deeper backbones. When equipped with stronger backbones, LSNet performs even better. This benefits from our proposed cross-IOU loss. It helps the LSNet to locate the landmarks with high accuracy, the rich global information contained in the landmarks further promote the cross-IOU loss to regress the landmarks more accurately. With the additional corner point verification (CPV)~\cite{chen2020reppoints} and multi-scale testing~\cite{zhang2020bridging}, LSNet achieves a box AP of $53.5 \%$, which outperforms all the anchor-free detectors as we know. 

\begin{table}[!tb]
\small
\centering
\renewcommand\tabcolsep{0.1cm} 
\begin{tabular}{c|c|ccc|ccc}
\hline
Loss & Box Style & AP & AP$_{50}$ & AP$_{75}$ & AP$_\mathrm{S}$ & AP$_\mathrm{M}$ &  AP$_\mathrm{L}$\\
\hline
\hline
GIOU             & rectangle & 34.6 & 54.7 & 36.6 & 19.3 & 38.2 & 44.4 \\
\hline                                    
Smooth-$\ell_1$  & rectangle & 33.8 & 54.5 & 36.0 & 18.6 & 37.1 & 43.8 \\
\hline 
Smooth-$\ell_1$  & extreme   & 31.5 & 50.1 & 34.0 & 16.8 & 34.3 & 41.4 \\
\hline 
\mycolor{Cross-IOU} & \mycolor{extreme}   & \mycolor{34.3} & \mycolor{54.9} & \mycolor{36.5} & \mycolor{19.6} & \mycolor{37.6} & \mycolor{44.7} \\
\hline 
\end{tabular}
\caption{The bounding box AP ($\%$) with different experimental settings. The cross-IOU loss achieves a high score even when regressing the extreme bounding box which is more difficult. Note that the GIOU loss cannot regress a non-rectangle bounding box.}
\label{tab:loss_contrast}
\vspace{-2ex}
\end{table}

\begin{table}[tb]
\small
\centering
\renewcommand\tabcolsep{0.1cm} 
\begin{tabular}{c|ccc|cccccc}
\hline
Loss & PA & PE & PP & AP & AP$_{50}$ & AP$_{75}$ & AP$_\mathrm{S}$ & AP$_\mathrm{M}$ & AP$_\mathrm{L}$\\
\hline
\hline
\multirow{3}{*}{Cross-IOU} & \checkmark &            &            & 34.3 & 54.9 & 36.5 & 19.6 & 37.6 & 44.7\\
                           & \checkmark & \checkmark &            & 35.5 & 54.8 & 38.3 & 20.0 & 39.2 & 45.3\\
                           & \checkmark &            & \checkmark &\textbf{36.2}&\textbf{55.4}&\textbf{38.9}
                                                                  &\textbf{19.8}&\textbf{39.8}&\textbf{46.3}\\
\hline
\end{tabular}
\caption{The detection accuracy ($\%$) of using different features. PA, PD, and PP denote using the anchor point features along, anchor point features with the single-scale extreme point features, and with the pyramid extreme point features, respectively.}
\label{tab:feature_contrast}
\vspace{-1ex}
\end{table}

\begin{table*}[!t]
	\small
	\begin{center}
		\resizebox{0.7\textwidth}{!}{
			\begin{tabular}{l|c|c|cccccc}
				\hline
				Method & Backbone & Epoch & AP & AP$_{50}$ & AP$_{75}$ & AP$_\mathrm{S}$ & AP$_\mathrm{M}$ & AP$_\mathrm{L}$\\
				\hline
				\hline
				\textbf{Pixel-based:} & & & & & & & &\\
				YOLACT~\cite{bolya2019yolact}  & R-101 & 48 &  31.2  & 50.6 & 32.8 & 12.1 & 33.3 & 47.1 \\
				TensorMask~\cite{chen2019tensormask} & R-101 & 72 & 37.1 & 59.3 & 39.4 & 17.1 & 39.1 & 51.6 \\
				Mask R-CNN~\cite{he2017mask} & X-101-32x4d & 12 & 37.1 & 60.0 & 39.4 & 16.9 & 39.9 & 53.5\\
				HTC~\cite{chen2019hybrid} & X-101-64x4d & 20 & 41.2 & 63.9 & 44.7 & 22.8 & 43.9 & 54.6\\
				DetectoRS~\cite{qiao2020detectors}~$\dagger$ & X-101-64x4d & 40 & 48.5 & 72.0 & 53.3 & 31.6 & 50.9 & 61.5 \\
				\hline
				\hline
				\textbf{Contour-based:} & & & & & & & &\\
				ExtremeNet~\cite{zhou2019bottom}  & HG-104 & 100 &  18.9  & 44.5 & 13.7 & 10.4 & 20.4 & 28.3 \\
				DeepSnake~\cite{peng2020deep} & DLA-34~\cite{yu2018deep} & 120 & 30.3  & - & - & -  & -  & - \\
				PolarMask~\cite{xie2020polarmask} & X-101-64x4d-DCN & 24 & 36.2  &59.4 &37.7 &17.8  &37.7  &51.5 \\
				\hline
				\mycolor{\textbf{LSNet}} & \mycolor{X-101-64x4d-DCN} & \mycolor{30} & \mycolor{37.6} & \mycolor{64.0} & \mycolor{38.3} & \mycolor{22.1} & \mycolor{39.9} & \mycolor{49.1}\\
				\mycolor{\textbf{LSNet}} & \mycolor{R2-101-DCN} & \mycolor{30} & \mycolor{38.0} & \mycolor{64.6} & \mycolor{39.0} & \mycolor{22.4} & \mycolor{40.6} & \mycolor{49.2}\\
				\mycolor{\textbf{LSNet}~$\dagger$} & \mycolor{X-101-64x4d-DCN} & \mycolor{30} & \mycolor{39.7} & \mycolor{65.5} & \mycolor{41.3} & \mycolor{25.5} & \mycolor{41.3} & 
				\mycolor{50.4}\\
				\mycolor{\textbf{LSNet}~$\dagger$} & \mycolor{R2-101-DCN} & \mycolor{30} & \mycolor{\textbf{40.2}} & \mycolor{\textbf{66.2}} & \mycolor{\textbf{42.1}} & 
				\mycolor{\textbf{25.8}} & \mycolor{\textbf{42.2}} & \mycolor{\textbf{51.0}}\\
				\hline
		\end{tabular}}
	\end{center}
	\caption{Comparison of LSNet to the sate-of-the-art methods in instance segmentation task on the COCO \textit{test-dev} set. Our LSNet achieves the state-of-the-art accuracy for contour-based instance segmentation. `R': ResNet~\cite{he2016deep}, `X': ResNeXt~\cite{xie2017aggregated}, `HG': Hourglass~\cite{newell2016stacked}, `R2':Res2Net~\cite{gao2019res2net}, `$\dagger$': multi-scale testing~\cite{zhang2020bridging}}
	\label{tab:segmentation}
\end{table*}

\noindent \textbf{Cross-IOU Loss for Vector Regression.} To evaluate the performance of cross-IOU loss, we design four contrast experiments on the MS COCO~\cite{lin2014microsoft} validation set, which are (i) the GIOU loss ~\cite{rezatofighi2019generalized} (a variant of the IOU loss) for rectangle bounding box regression, (ii) the smooth-$\ell_1$ loss for rectangle bounding box regression, (iii) the smooth-$\ell_1$ loss for extreme bounding box regression, and (iv) the cross-IOU loss for extreme bounding box regression, respectively. All the experiments are done in the first stage in our framework (shown in Figure~\ref{fig:architecture}) with ResNet-50~\cite{he2016deep} as the backbone, and we train the model for each experiment for $12$ epochs. Table~\ref{tab:loss_contrast} summarizes the results. We can see that the smooth-$\ell_1$ loss reports an AP of $33.8 \%$ and $31.5\%$ when regressing the rectangle bounding boxes and the the extreme bounding boxes, respectively. This reveals that it is more difficult to regress an angled vector than a straight vector. By contrast, the cross-IOU loss performs much better than the smooth-$\ell_1$ loss and even produces competitive results with the IOU loss for the rectangle bounding box. Although the GIOU loss at present still performs better than the cross-IOU loss, the cross-IOU loss allows the framework regressing the location of the landmarks, thus we could extract the discriminative information around the landmarks to enhance recognition. We will show in the next section that the combination of the cross-IOU loss and landmark feature extraction significantly boosts recognition accuracy.

\begin{figure}[!tb]
  \centering 
  \subfigure[]{ 
    \includegraphics[width = 0.22\textwidth]{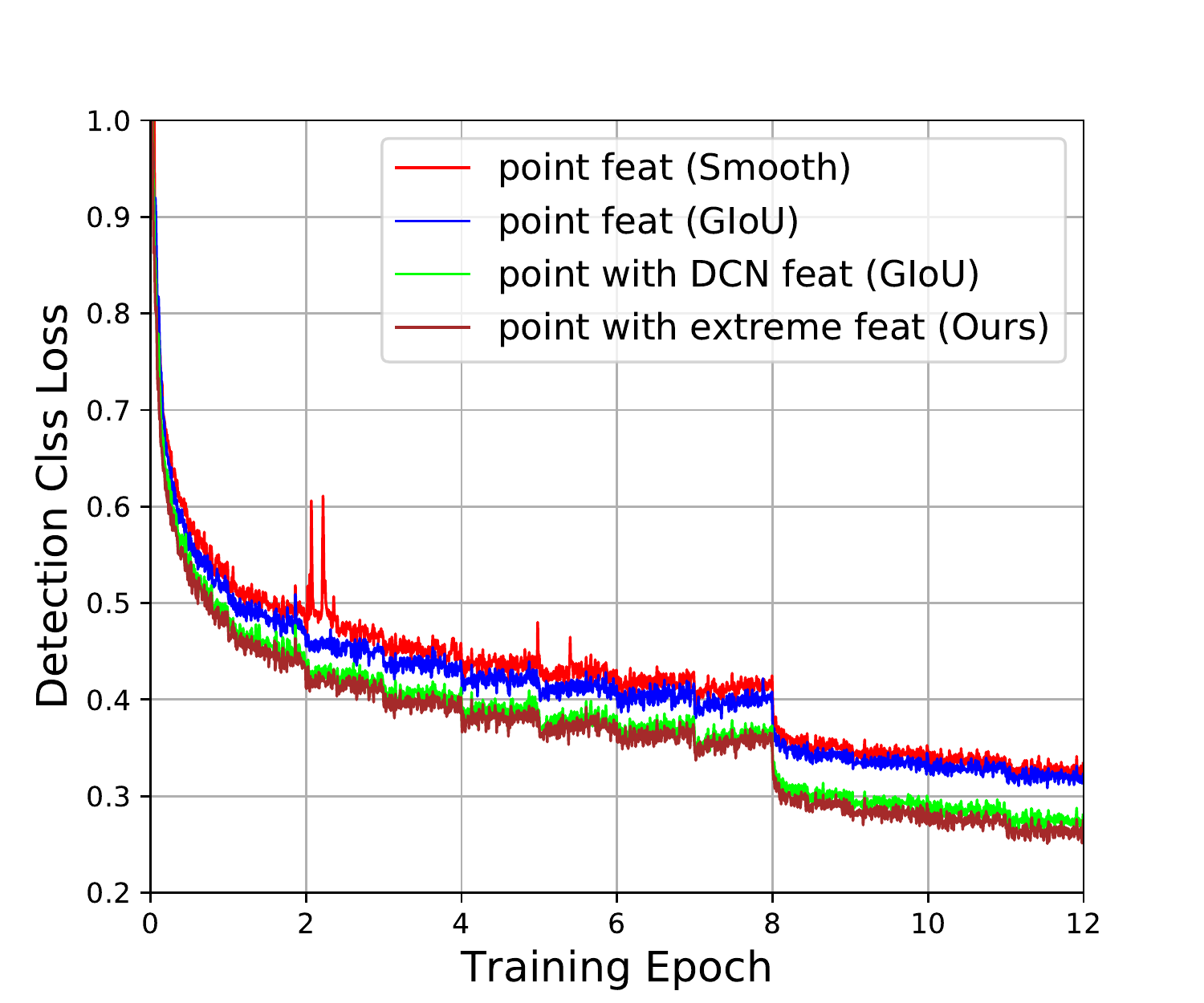}
    \label{fig:cls_loss} 
  } 
  \subfigure[]{ 
    \includegraphics[width = 0.225\textwidth]{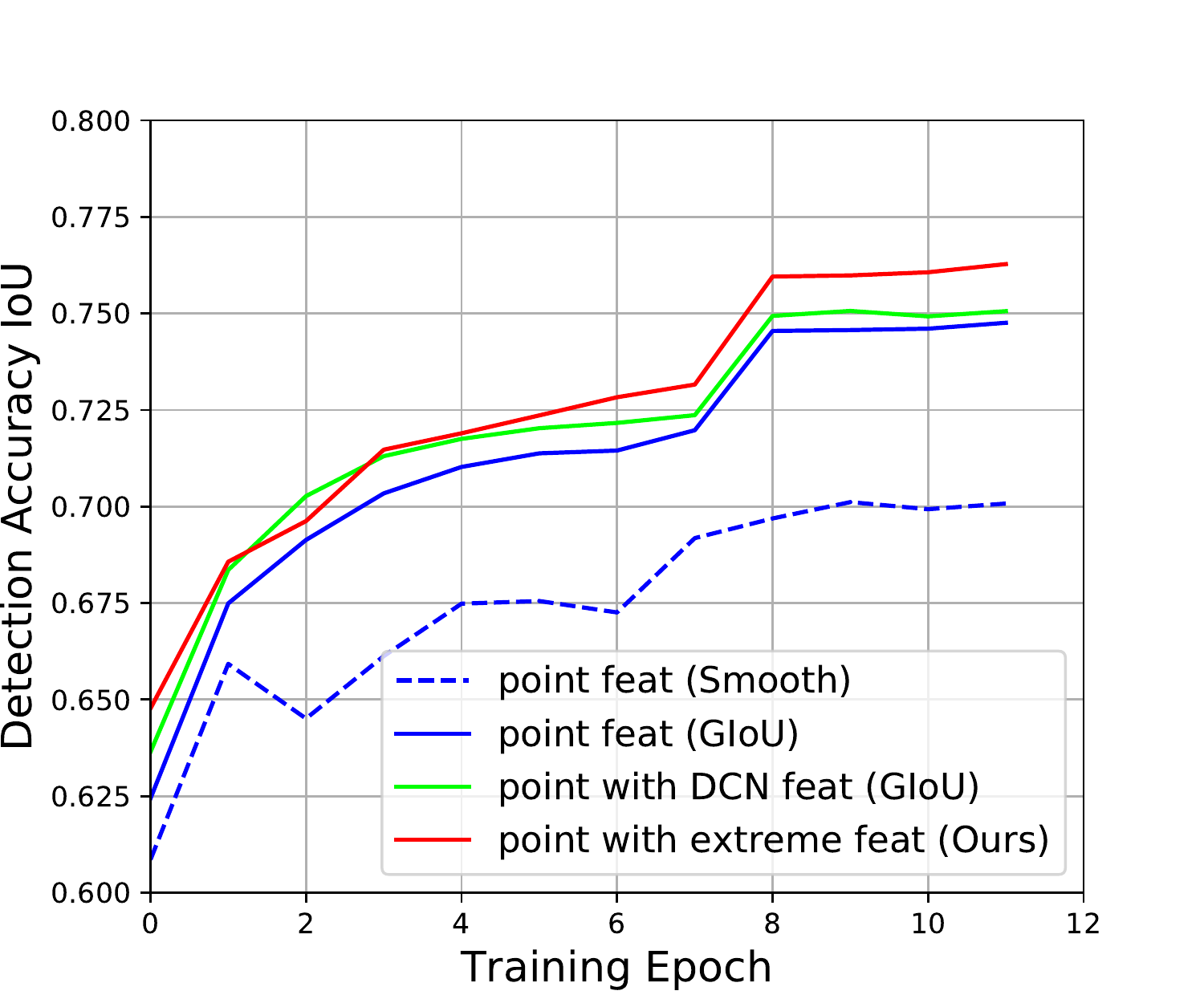}
    \label{fig:iou_acc} 
  }
  \caption{The detection classification loss and average IOU with respect to the number of elapsed epochs. `DCN feat', `point feat' and `extreme feat' denote the features adaptively learned by the DCN kernel, the features at the anchor points and extreme points, respectively. 'Smooth', 'GIOU' and 'Ours' denote the smooth-$\ell_1$ loss, the GIOU loss~\cite{rezatofighi2019generalized}, and the cross-IOU loss, respectively.}
  \label{fig:loss_acc}
\end{figure}

\noindent \textbf{Landmark Features Improve Precision.} The landmarks (in particular, the extreme points in the detection task) are often related to discriminative appearance features, which may benefit visual recognition. To confirm this, we investigate different settings by using the anchor point features alone and integrating the anchor point features with either DCN or extreme point features, respectively. We still report the detection accuracy with all other settings remaining the same as in the previous experiments (studying the cross IOU loss). The results are shown in Figure~\ref{fig:loss_acc}. For the smooth-$\ell_1$ loss and the GIOU loss, we regress the rectangle bounding boxes, and the DCN features are extracted by the adaptively learned DCN kernel; for the cross-IOU loss, we use two sets of vectors both of which regress the extreme bounding boxes -- the first set is trained to predict the extreme points and extract the extreme features, and we use the extreme features along with the anchor point features to train the second set from scratch. As shown in Figure~\ref{fig:loss_acc}, both the extreme and DCN features boost the classification accuracy. Recall that the prior experiments suggested the usefulness of the extreme features are useful for localization, combining the current results, we verify that the features around the landmarks are discriminative and thus benefit visual recognition.

\noindent \textbf{Pyramid DCN Improves Precision.} We further equip the LSNet with the pyramid DCN to extract the multi-scale features around the landmarks. Table~\ref{tab:feature_contrast} shows our method achieves an AP of $36.2\%$ with the features extracted by the Pyramid DCN, which outperforms the AP with single-scale features by a margin of $0.7\%$.

\begin{table*}[!tb]
\begin{minipage}{0.65\linewidth}
\small
\begin{center}
\renewcommand{\multirowsetup}{\centering}
\resizebox{0.9\textwidth}{!}{
\begin{tabular}{l|c|c|c|cccc}
\hline
Method & Backbone & Epoch & AP & AP$_{50}$ & AP$_{75}$ & AP$_\mathrm{M}$ & AP$_\mathrm{L}$\\
\hline
\hline
\textbf{Heatmap-based:} & & & & & & &\\
CenterNet-jd~\cite{zhou2019objects} & DLA-34 & 320 & 57.9& 84.7 & 63.1 & 52.5 & 67.4\\
OpenPose~\cite{cao2019openpose} & VGG-19 & - & 61.8 & 84.9 & 67.5 & 58.0 & 70.4\\
Pose-AE~\cite{newell2017associative} & HG & 300 & 62.8 & 84.6 & 69.2 & 57.5 & 70.6\\
CenterNet-jd~\cite{zhou2019objects} & HG104 & 150 & 63.0 & 86.8 & 69.6 & 58.9 & 70.4\\
Mask R-CNN~\cite{he2017mask} & R-50 & 28 & 63.1 & 87.3 & 68.7 & 57.8 & 71.4\\
PersonLab~\cite{papandreou2018personlab} & R-152 & $>$1000 & 66.5 & 85.5 & 71.3 & 62.3 & 70.0\\
HRNet~\cite{sun2019deep} & HRNet-W32 & 210 & 74.9 & 92.5 & 82.8 & 71.3 & 80.9\\
\hline
\hline
\textbf{Regression-based:} & & & & & & &\\
\multirow{2}{*}{CenterNet-reg~\cite{zhou2019objects}}
                                   & DLA-34 & 320 & 51.7& 81.4 & 55.2 & 44.6 & 63.0\\
                                   & HG-104 & 150 & 55.0 & 83.5 & 59.7 & 49.4 & 64.0\\
                        
\hline
\mycolor{\textbf{LSNet w/ obj-box}}          & \mycolor{X-101-64x4d-DCN} & \mycolor{60} & \mycolor{55.7} & \mycolor{81.3} & \mycolor{61.0} & \mycolor{52.9} & 
\mycolor{60.5}\\
\hline  
\mycolor{\textbf{LSNet w/ kps-box}}          & \mycolor{X-101-64x4d-DCN} & \mycolor{20} & \mycolor{59.0} & \mycolor{83.6} & \mycolor{65.2} & \mycolor{53.3} &
\mycolor{67.9}\\
\hline  
\end{tabular}}
\end{center}
\end{minipage}
\hfill
\begin{minipage}{.35\linewidth}
\vspace{-0.7ex}
\centerline{\includegraphics[width=2.8cm]{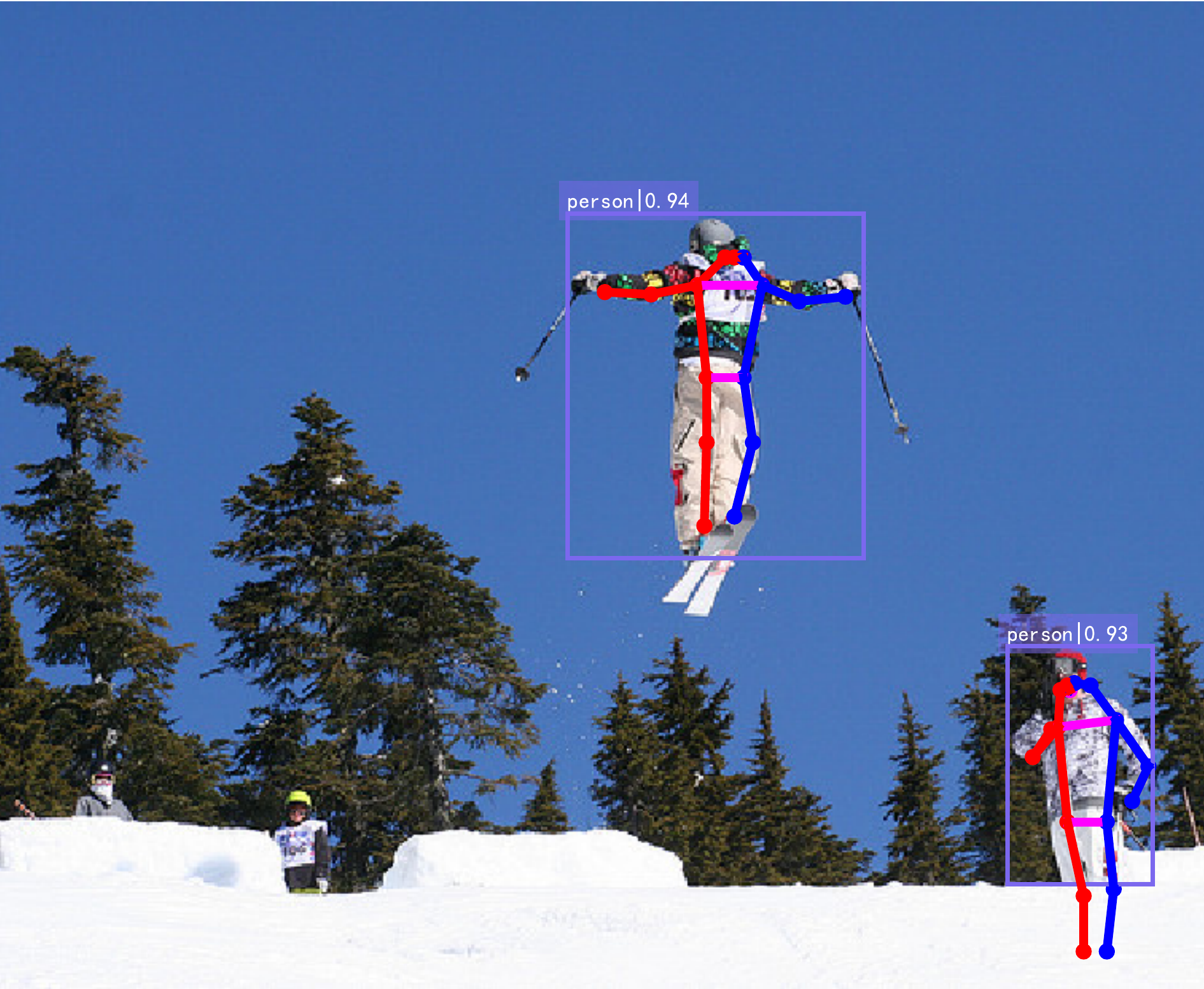}
            \includegraphics[width=2.8cm, height=2.3cm]{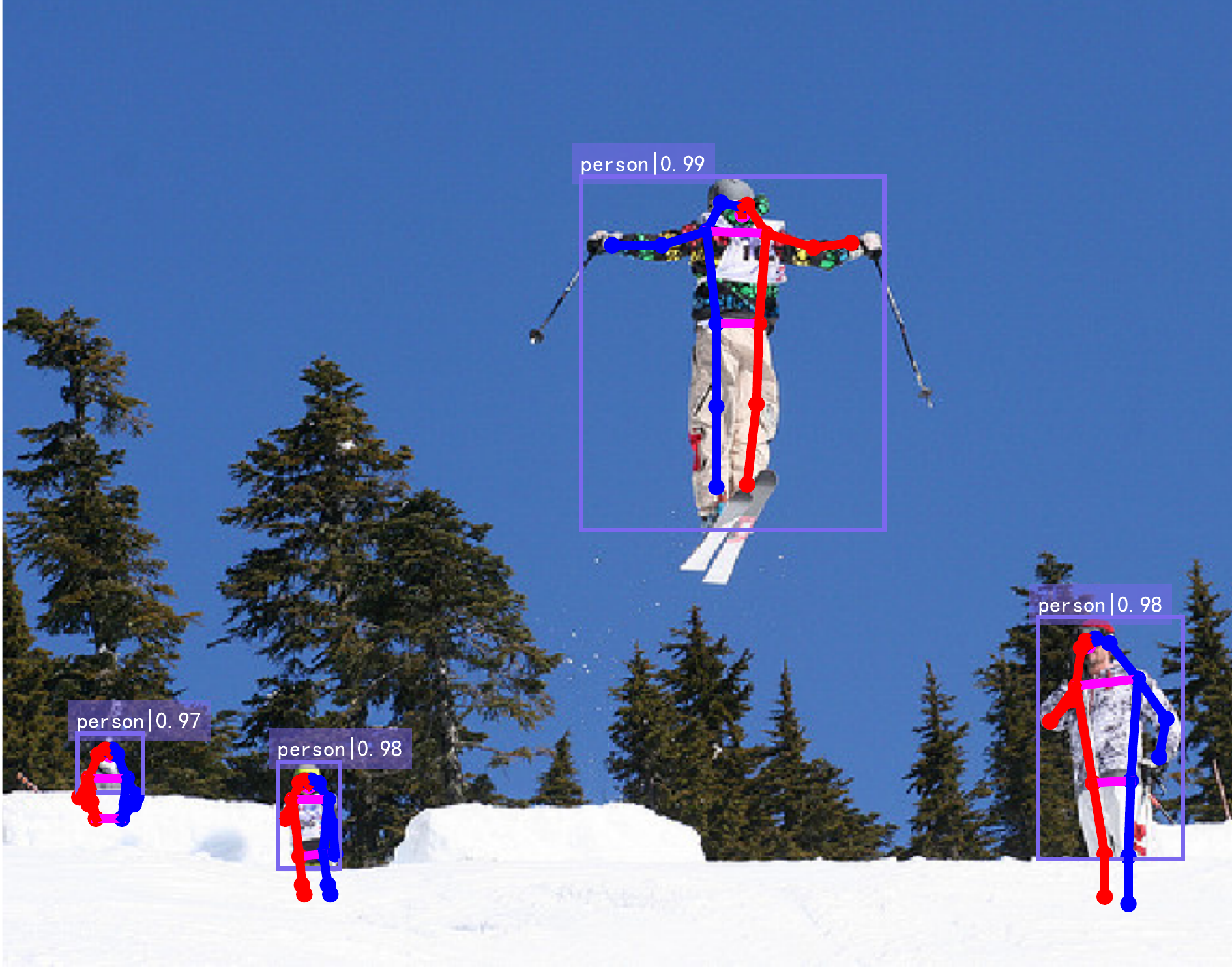}}

\centerline{\includegraphics[width=2.8cm]{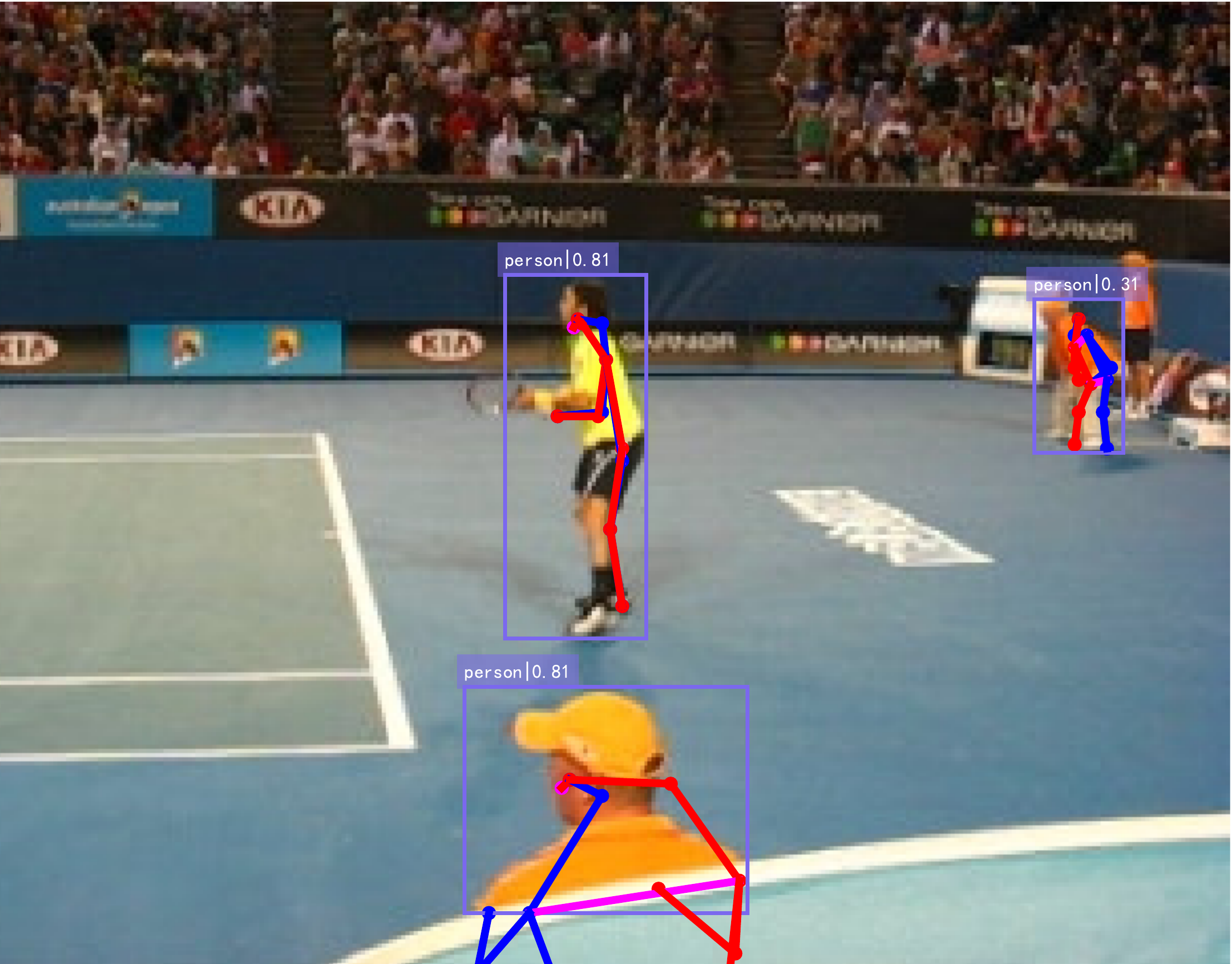}
            \includegraphics[width=2.8cm,height=2.22cm]{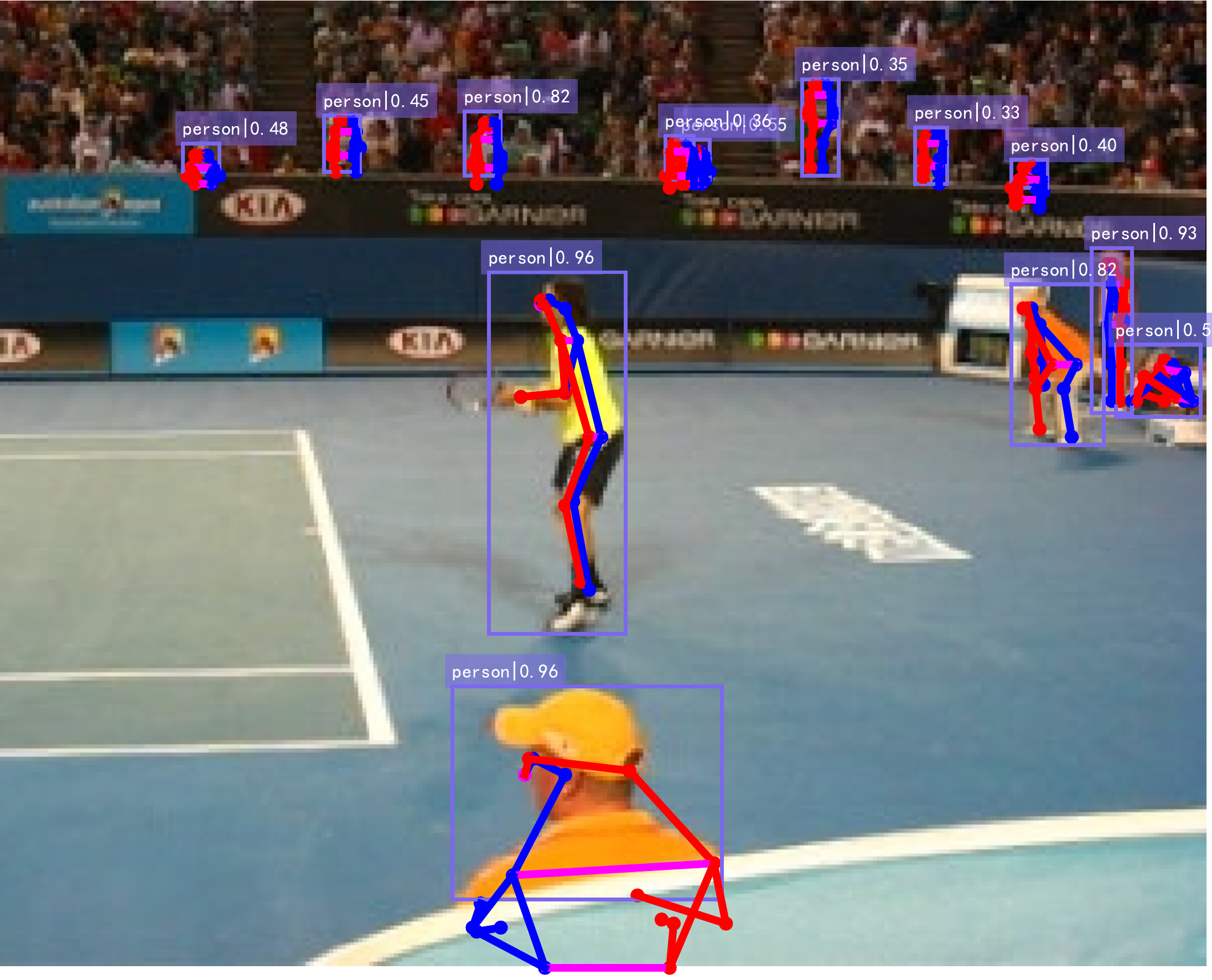}
             } 
\end{minipage}
\caption{Left: Comparison of LSNet to the sate-of-the-art methods in pose estimation task on the COCO \textit{test-dev} set. LSNet predict the keypoints by regression. `obj-box' and `kps-box' denote the object bounding boxes and the keypoint-boxes, respectively. For LSNet w/ kps-box, we fine-tune the model from the LSNet w/ kps-box for another $20$ epochs. Right: We compared with the CenterNet~\cite{zhou2019objects} to show that our LSNet w/ `obj-box' tends to predict more human pose of small scales, which are not annotated on the dataset. Only pose results with scores higher than $0.3$ are shown for both methods.}
\label{tab:pose}
\end{table*}

\begin{figure*}[tb]
  \centering 
  \subfigure{ 
    \includegraphics[height=0.12\textwidth,width=0.07\textheight]{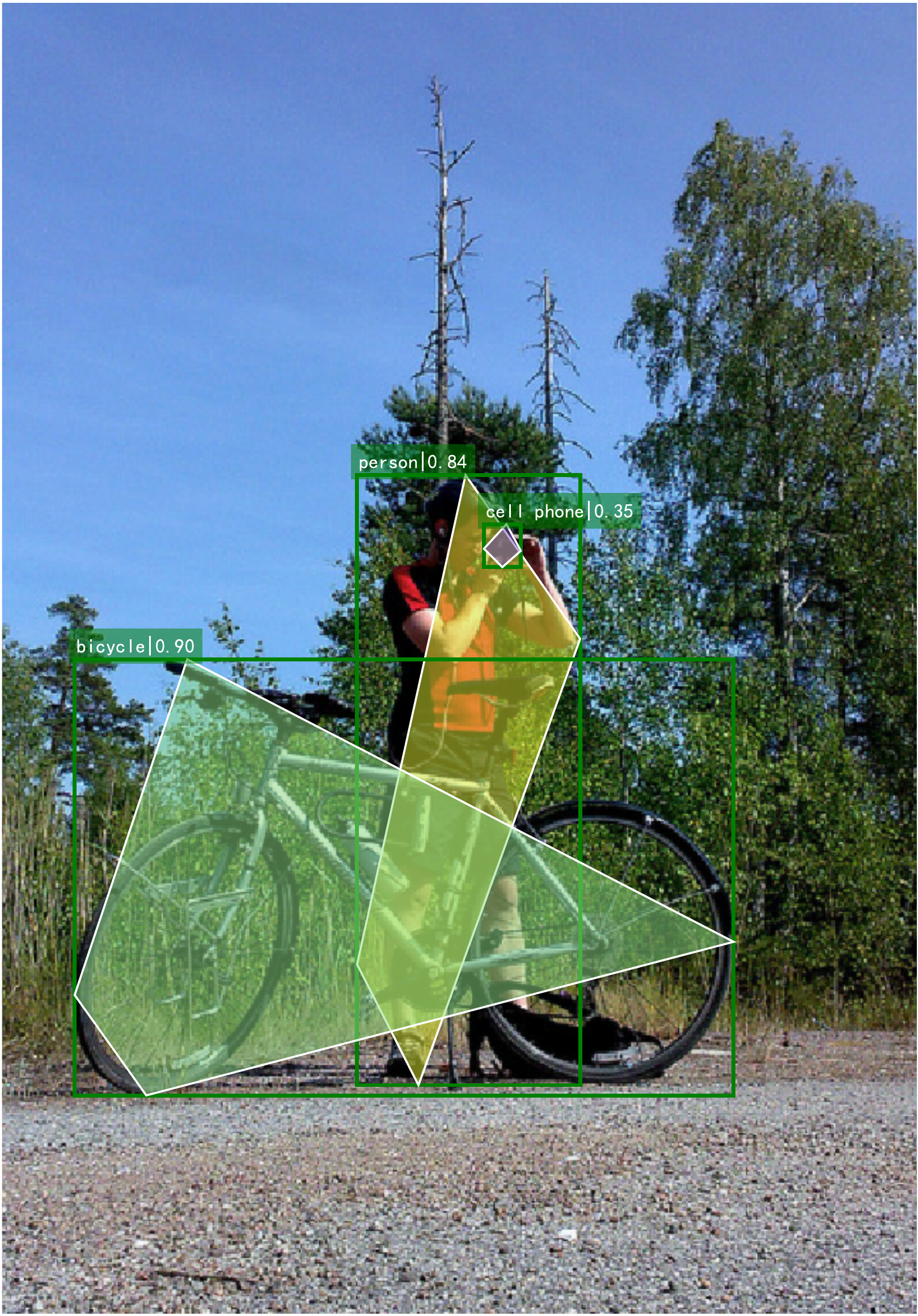}
    \label{fig8_1}
  }
  \subfigure{ 
    \includegraphics[height=0.12\textwidth,width=0.1\textheight]{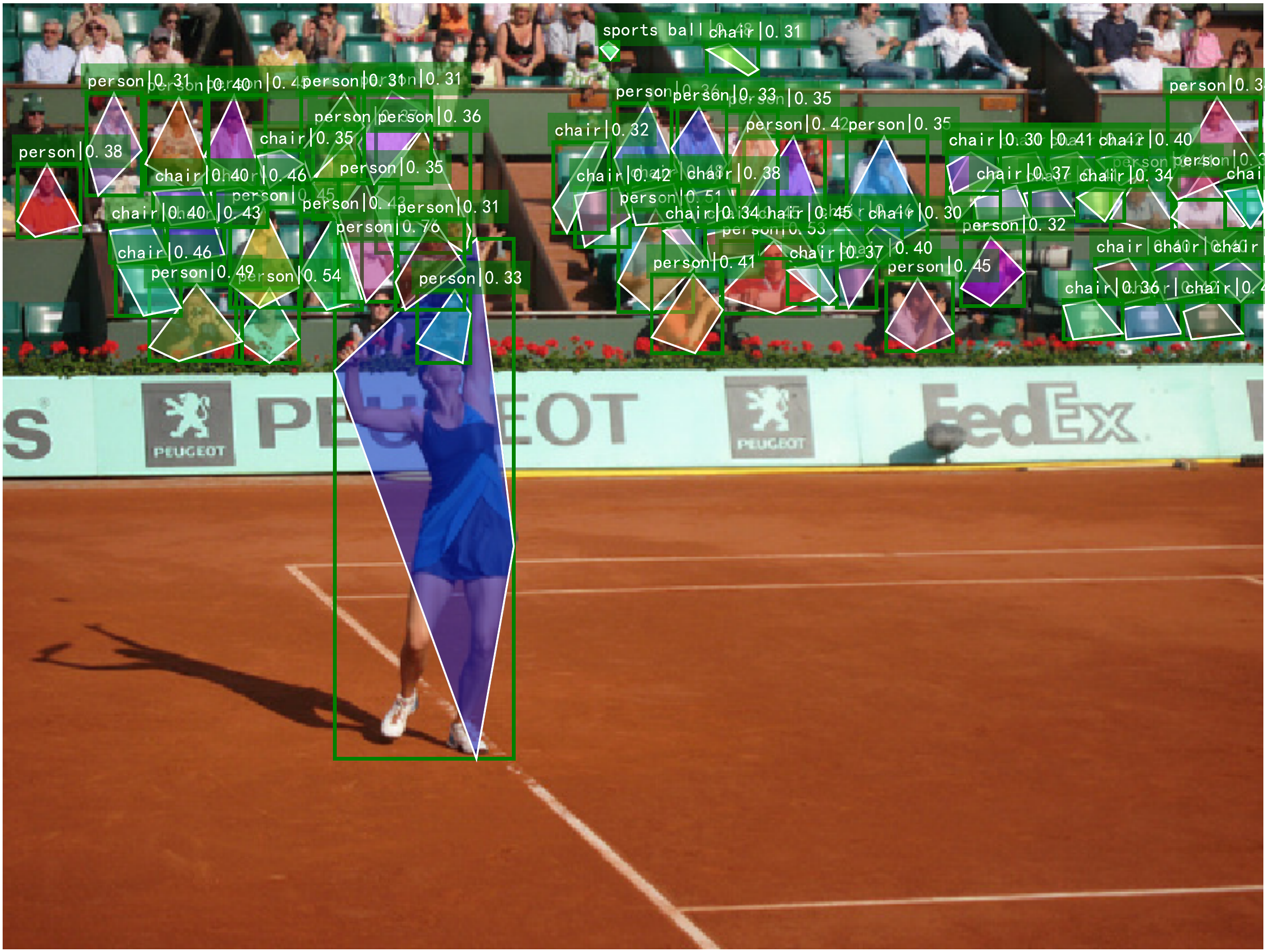}
    \label{fig8_2}
  }
  \subfigure{ 
    \includegraphics[height=0.12\textwidth,width=0.14\textheight]{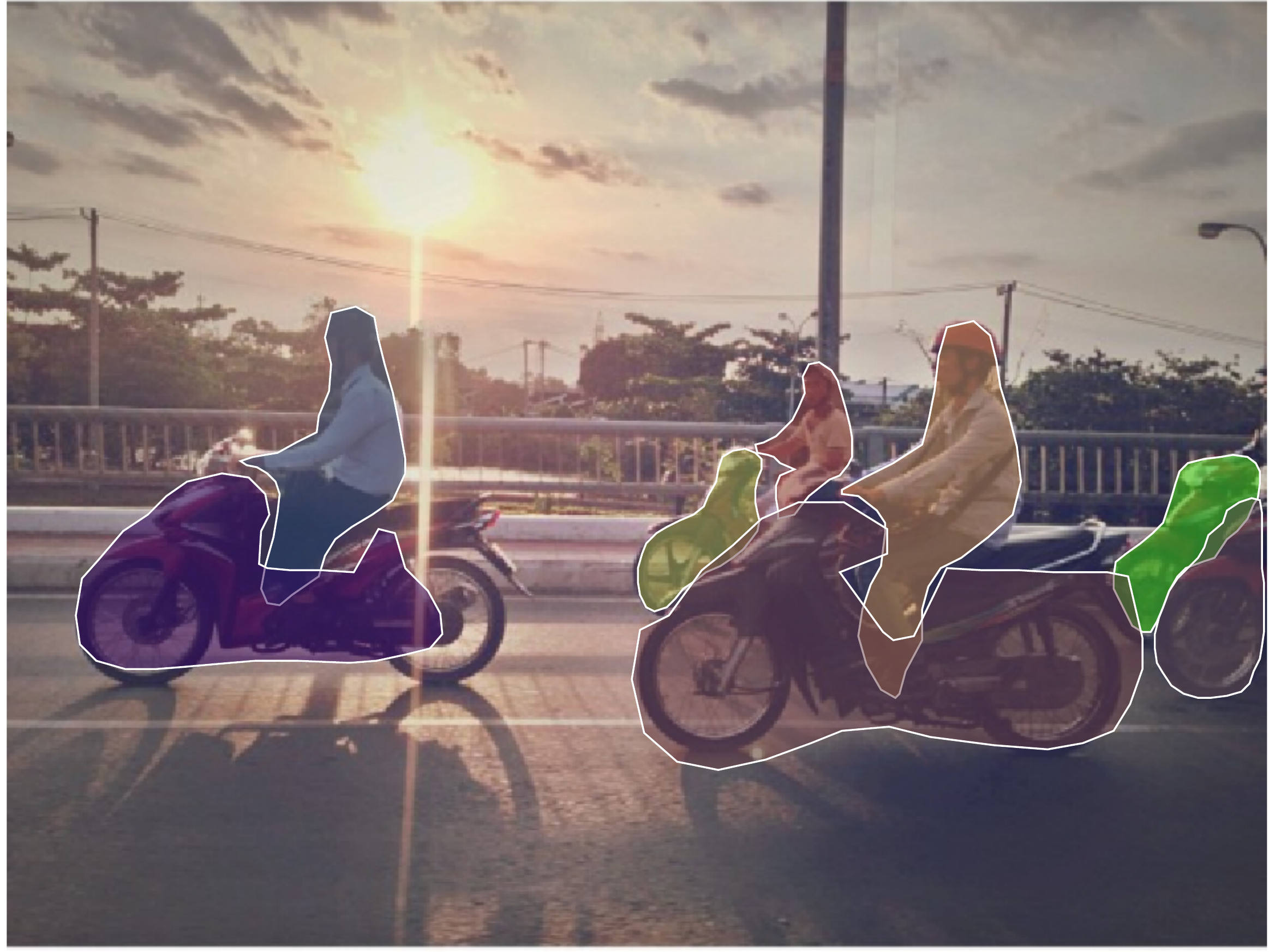}
    \label{fig8_3}
  } 
  \subfigure{ 
    \includegraphics[height=0.12\textwidth,width=0.14\textheight]{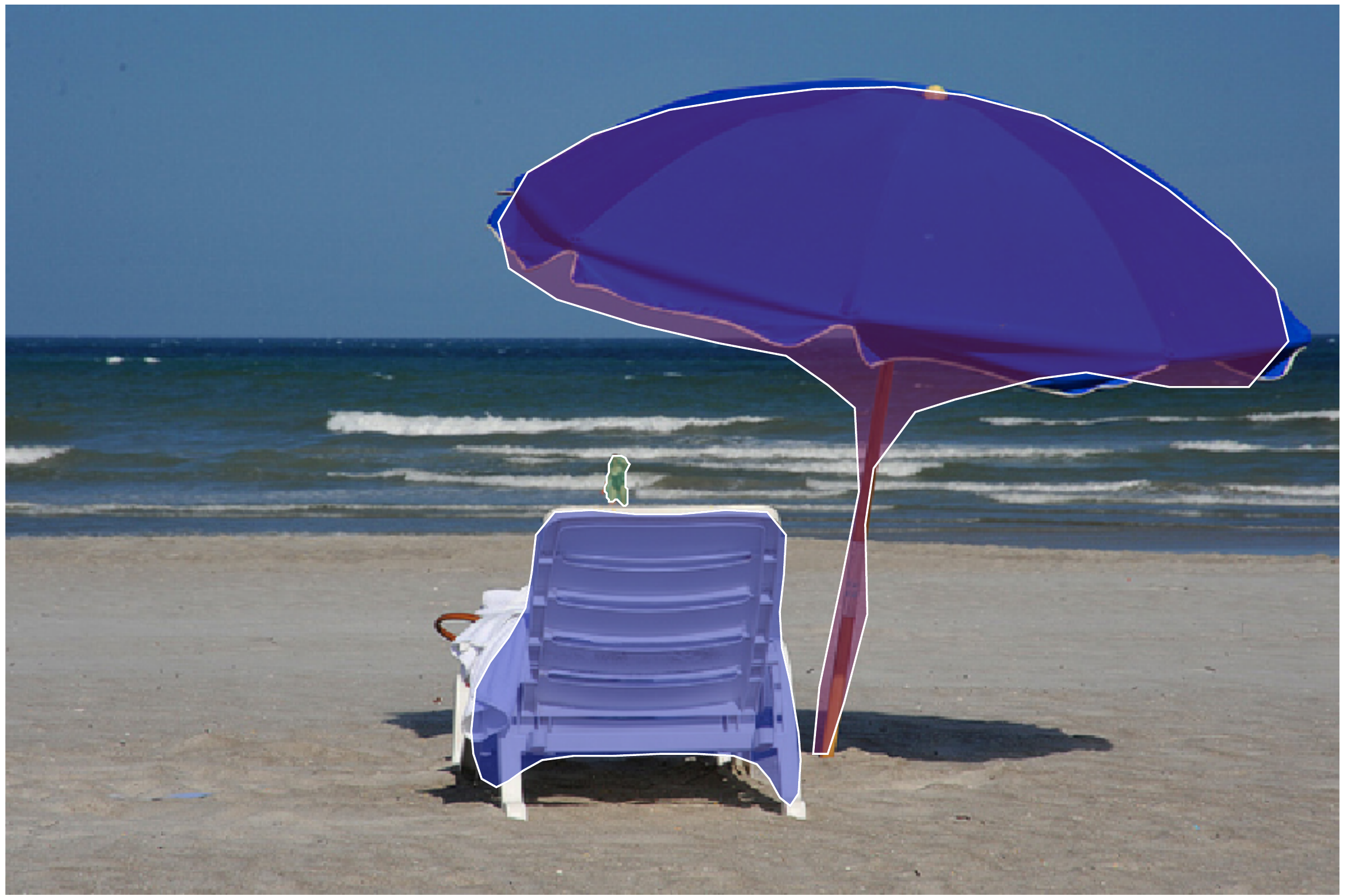}
    \label{fig8_4}
  } 
  \subfigure{ 
    \includegraphics[height=0.12\textwidth,width=0.1\textheight]{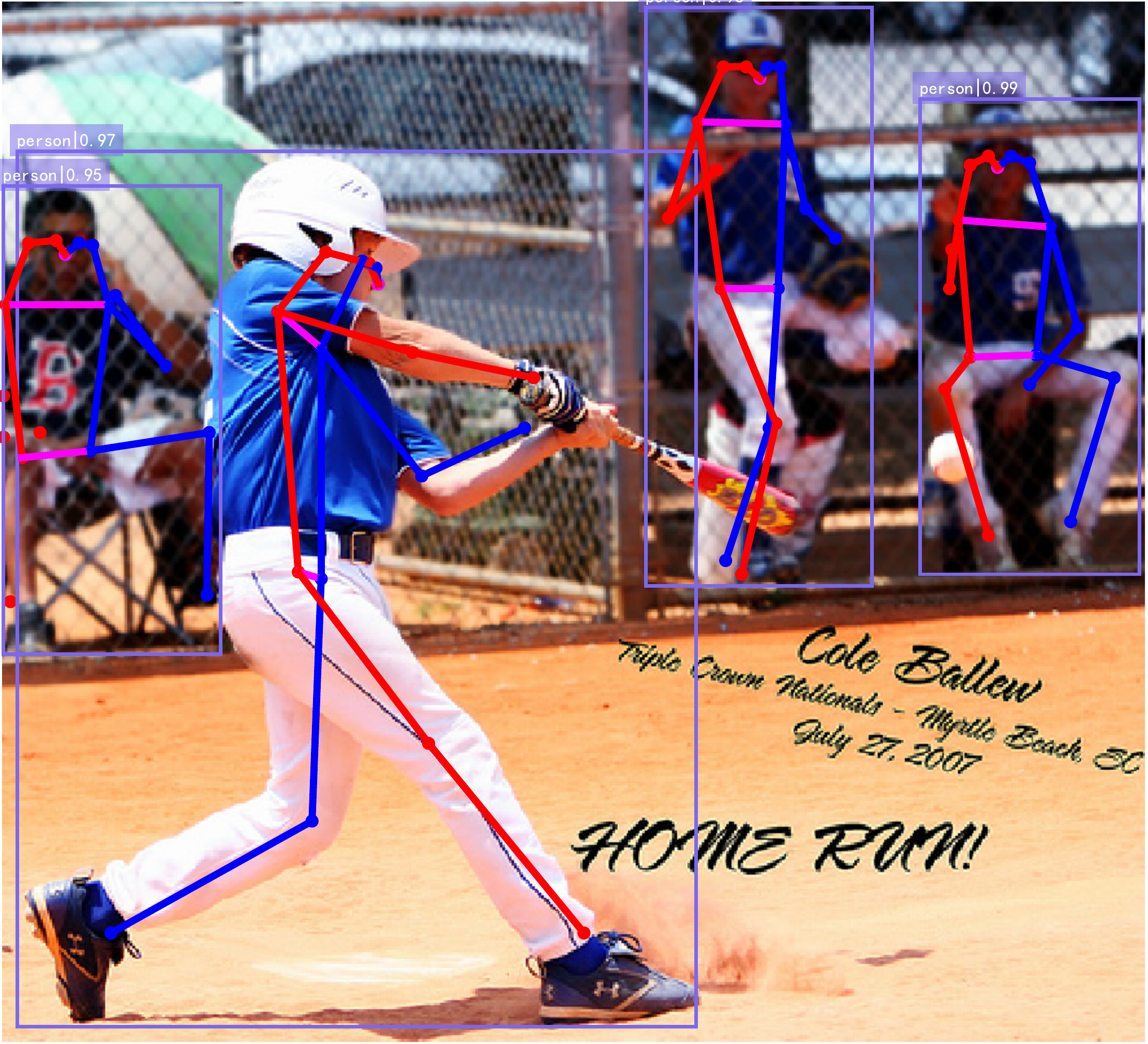}
    \label{fig8_5}
  } 
  \vspace{-1ex}
  \subfigure{ 
    \includegraphics[height=0.12\textwidth,width=0.15\textheight]{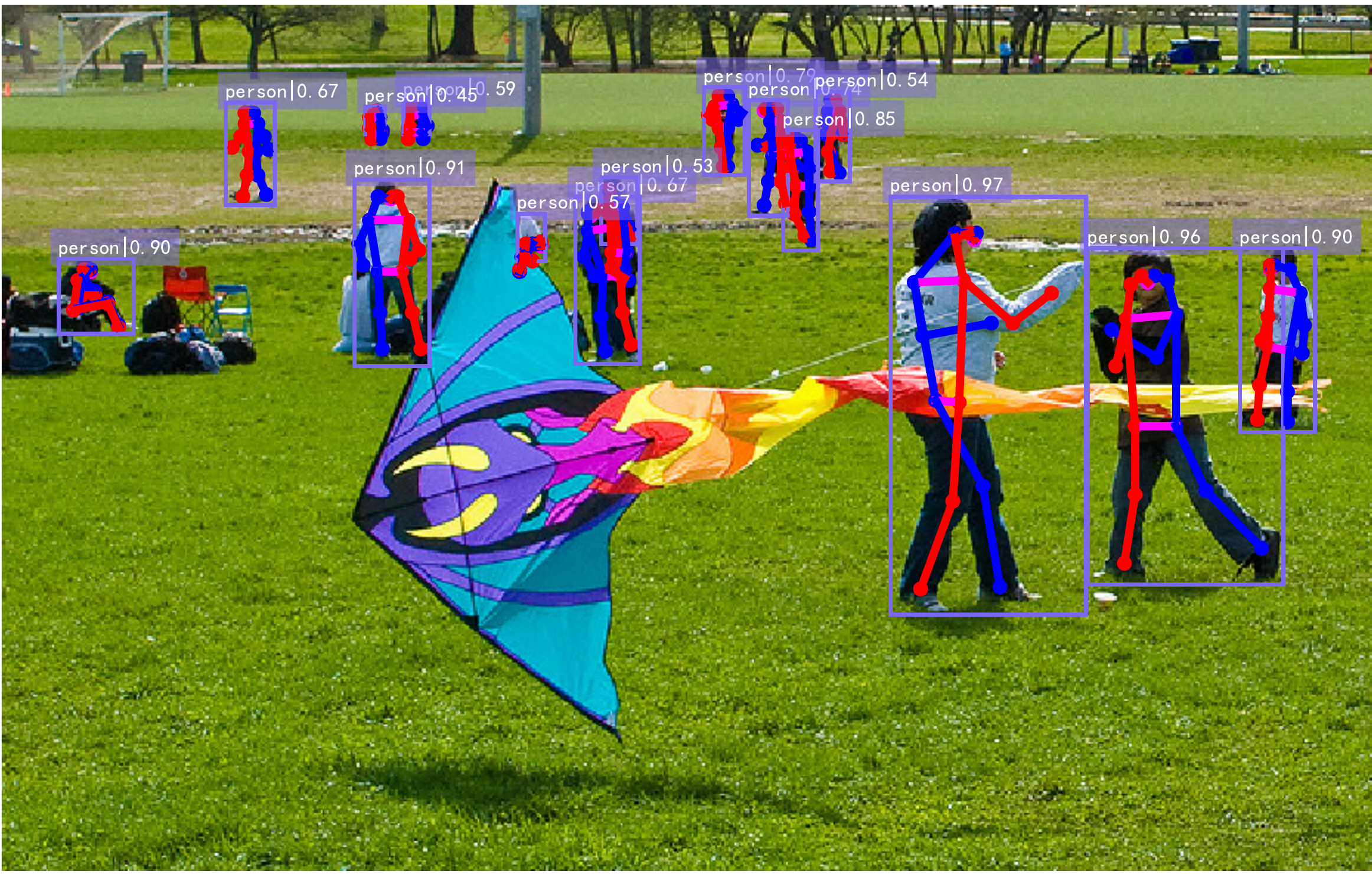}
    \label{fig8_6}
  }
  \subfigure{ 
    \includegraphics[height=0.12\textwidth,width=0.07\textheight]{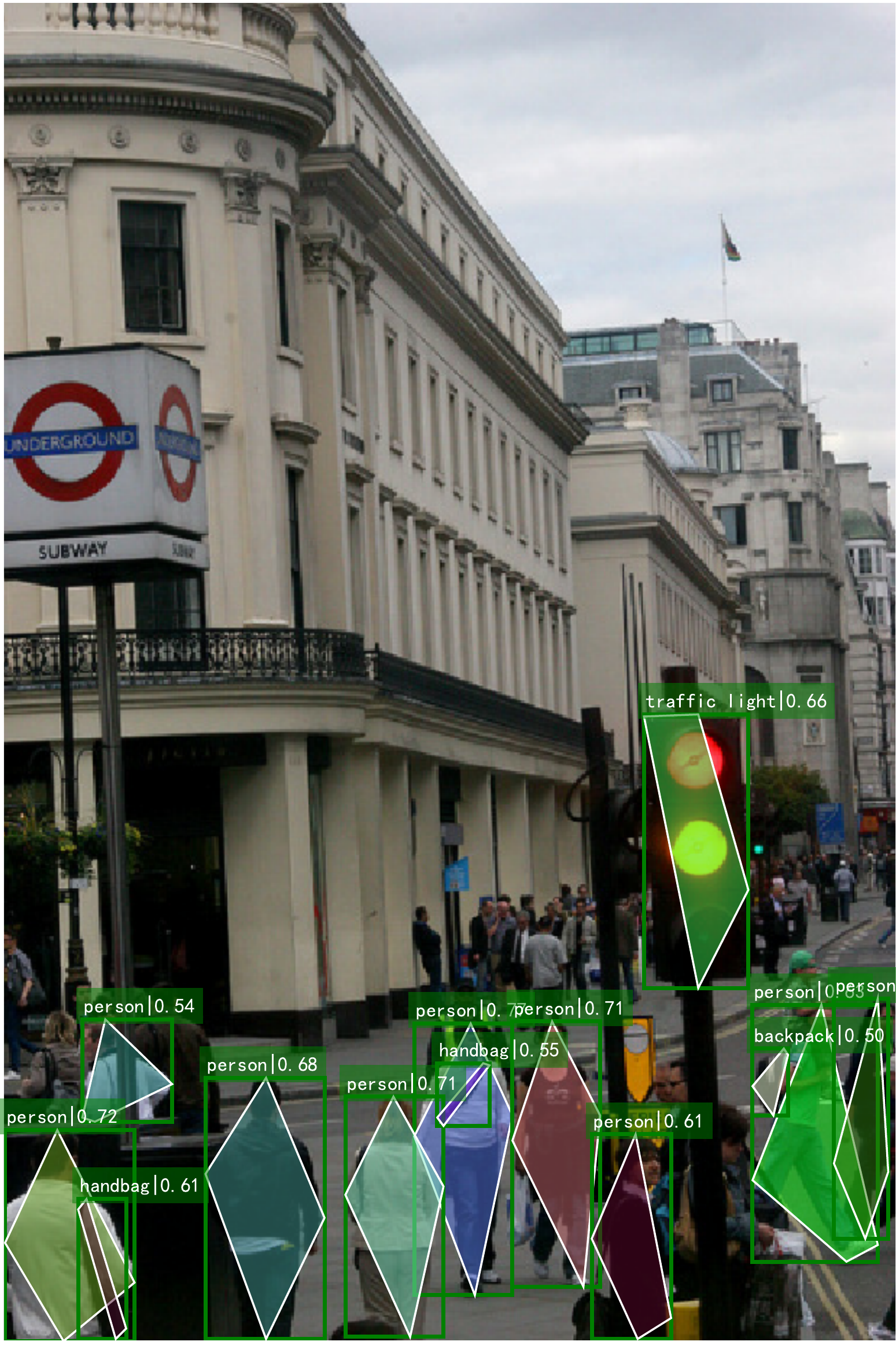}
    \label{fig8_1}
  }
  \subfigure{ 
    \includegraphics[height=0.12\textwidth,width=0.1\textheight]{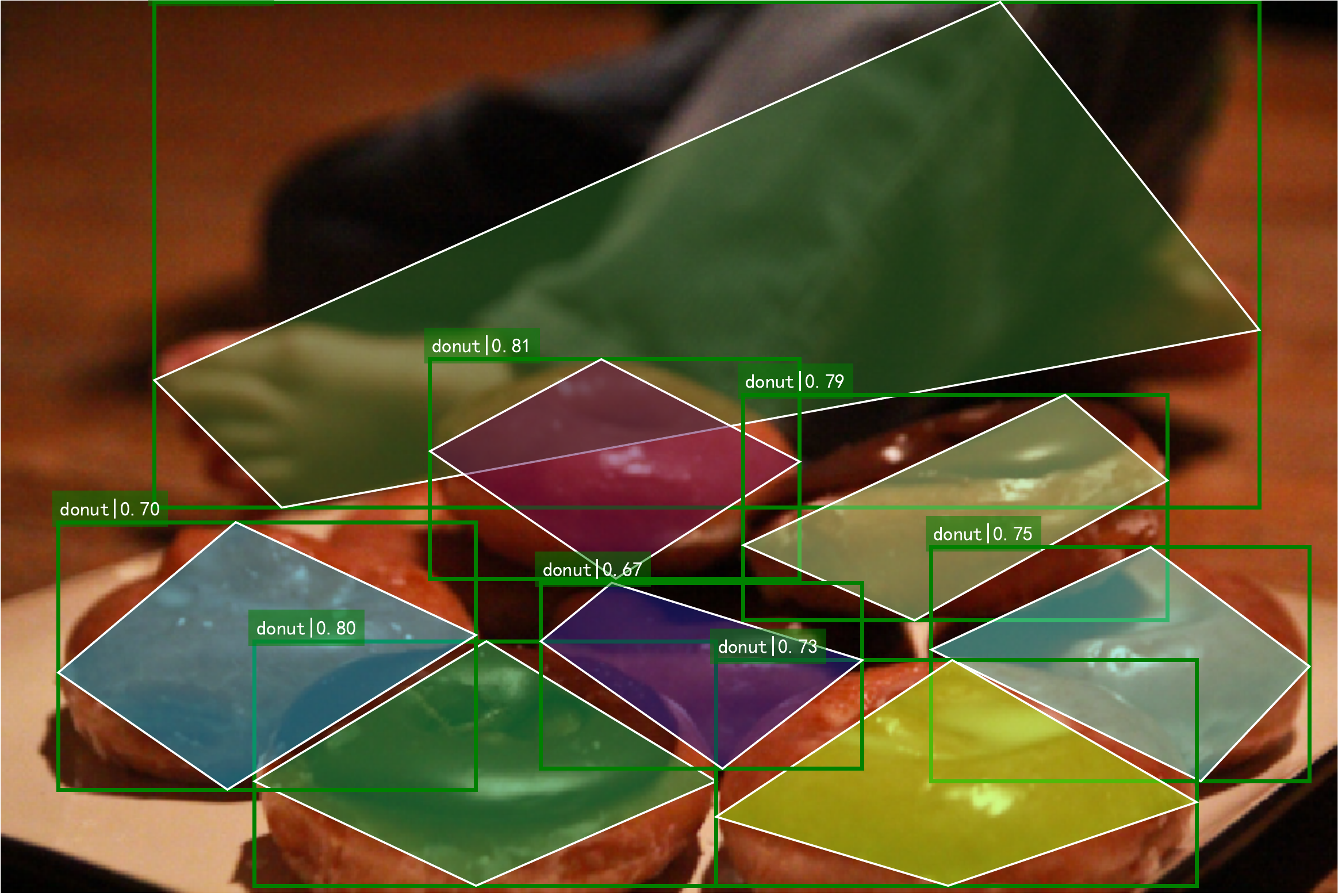}
    \label{fig8_2}
  }
  \subfigure{ 
    \includegraphics[height=0.12\textwidth,width=0.14\textheight]{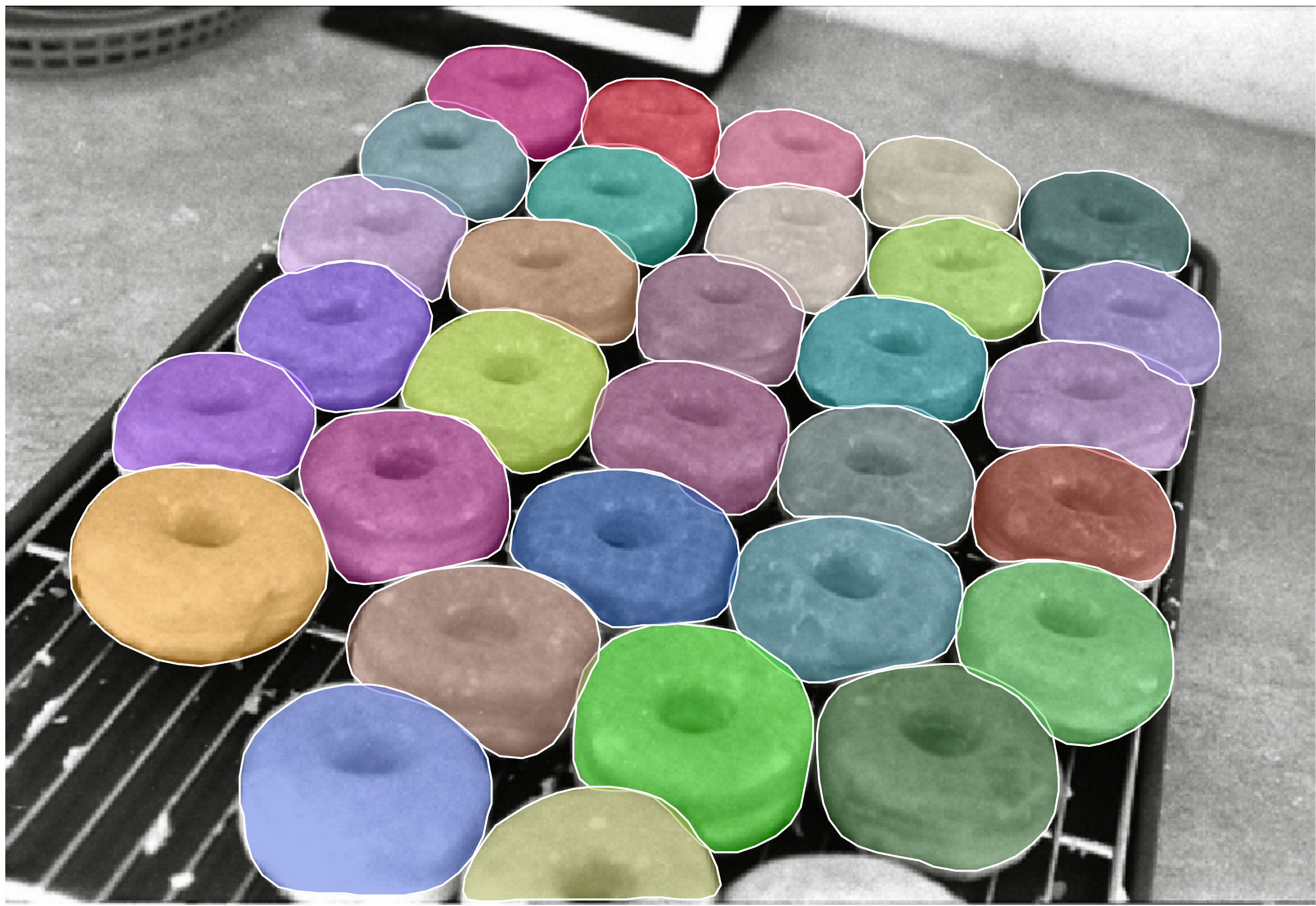}
    \label{fig8_3}
  } 
  \subfigure{ 
    \includegraphics[height=0.12\textwidth,width=0.14\textheight]{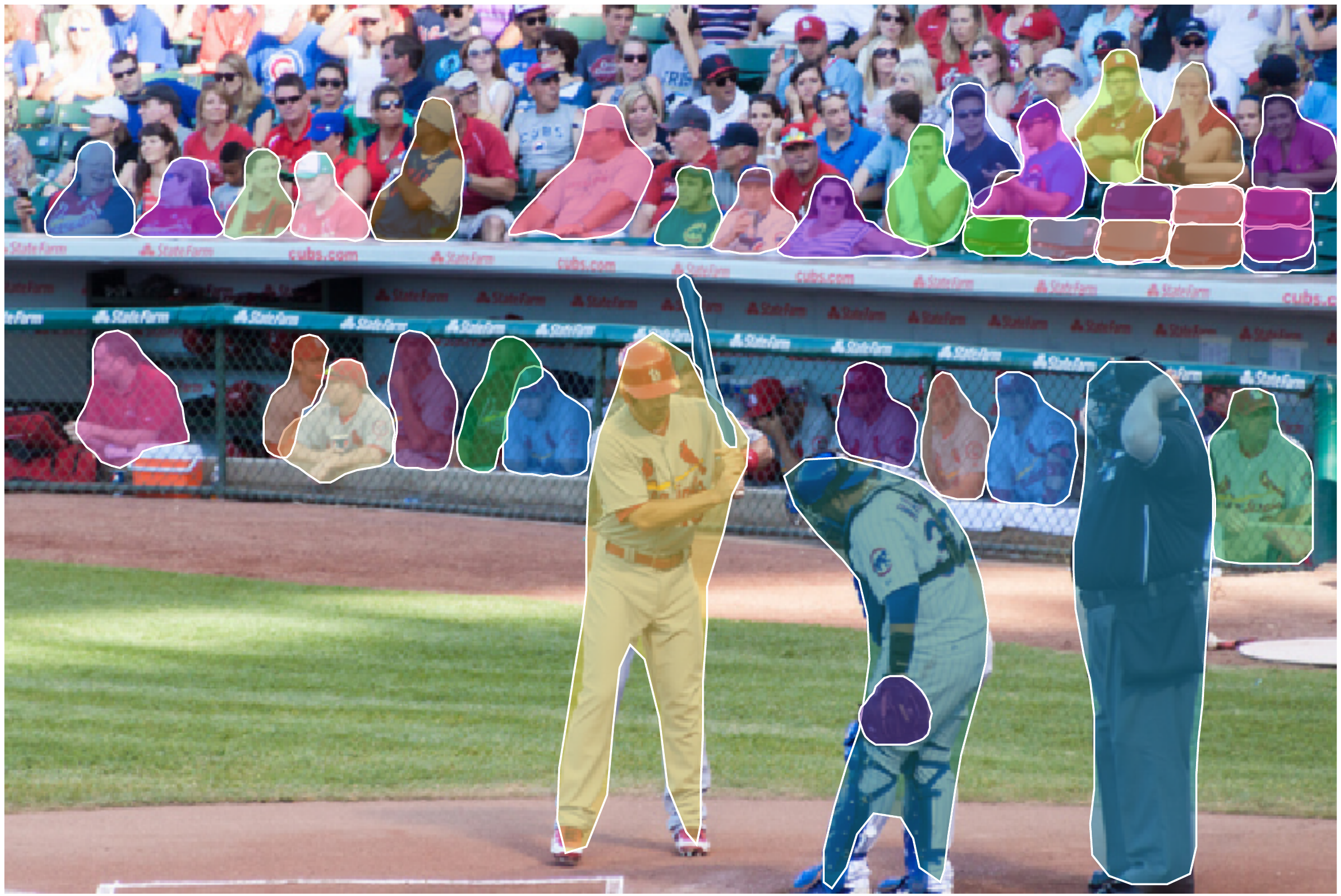}
    \label{fig8_4}
  } 
  \subfigure{ 
    \includegraphics[height=0.12\textwidth,width=0.1\textheight]{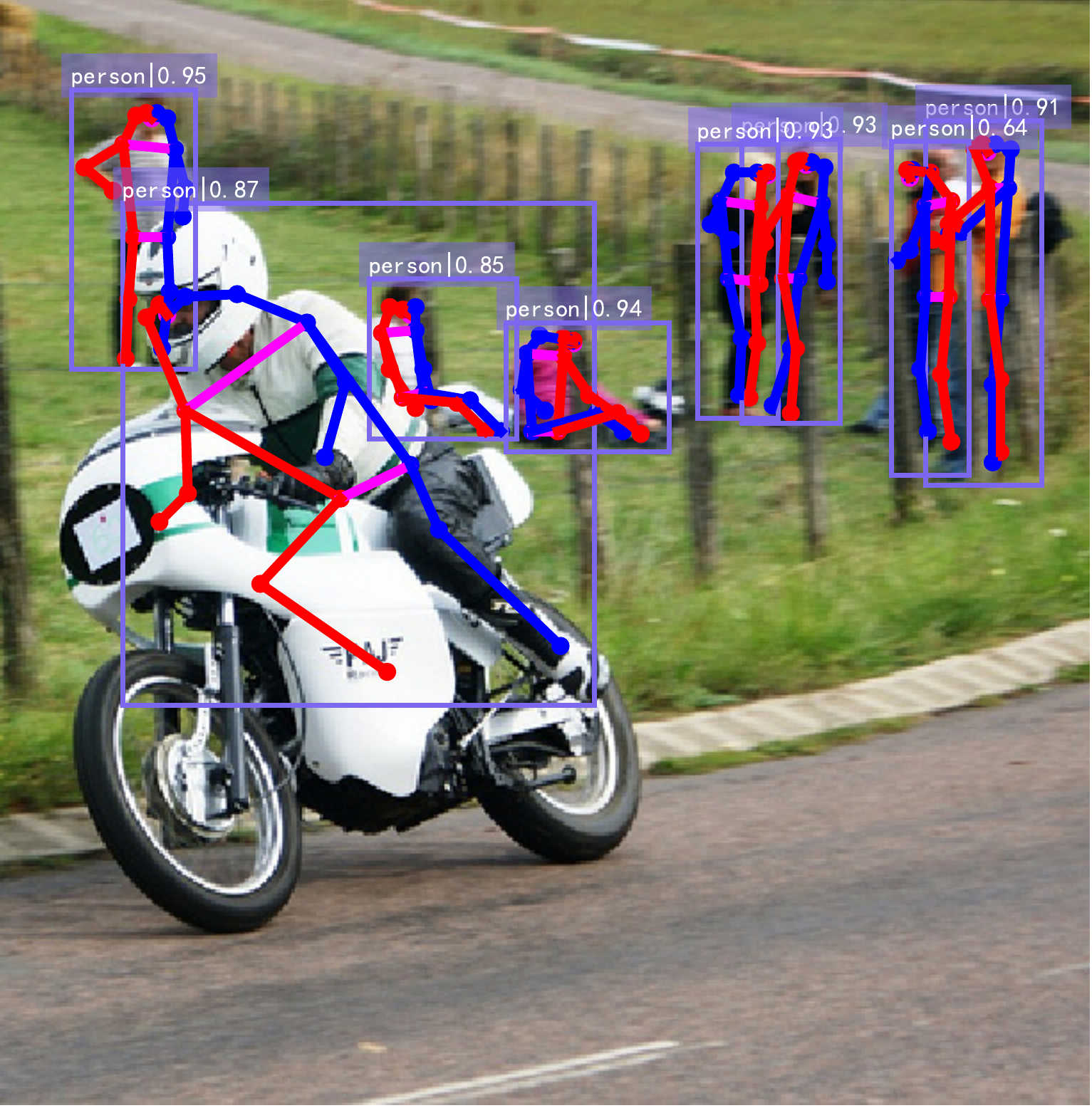}
    \label{fig8_5}
  } 
  \subfigure{ 
    \includegraphics[height=0.12\textwidth,width=0.15\textheight]{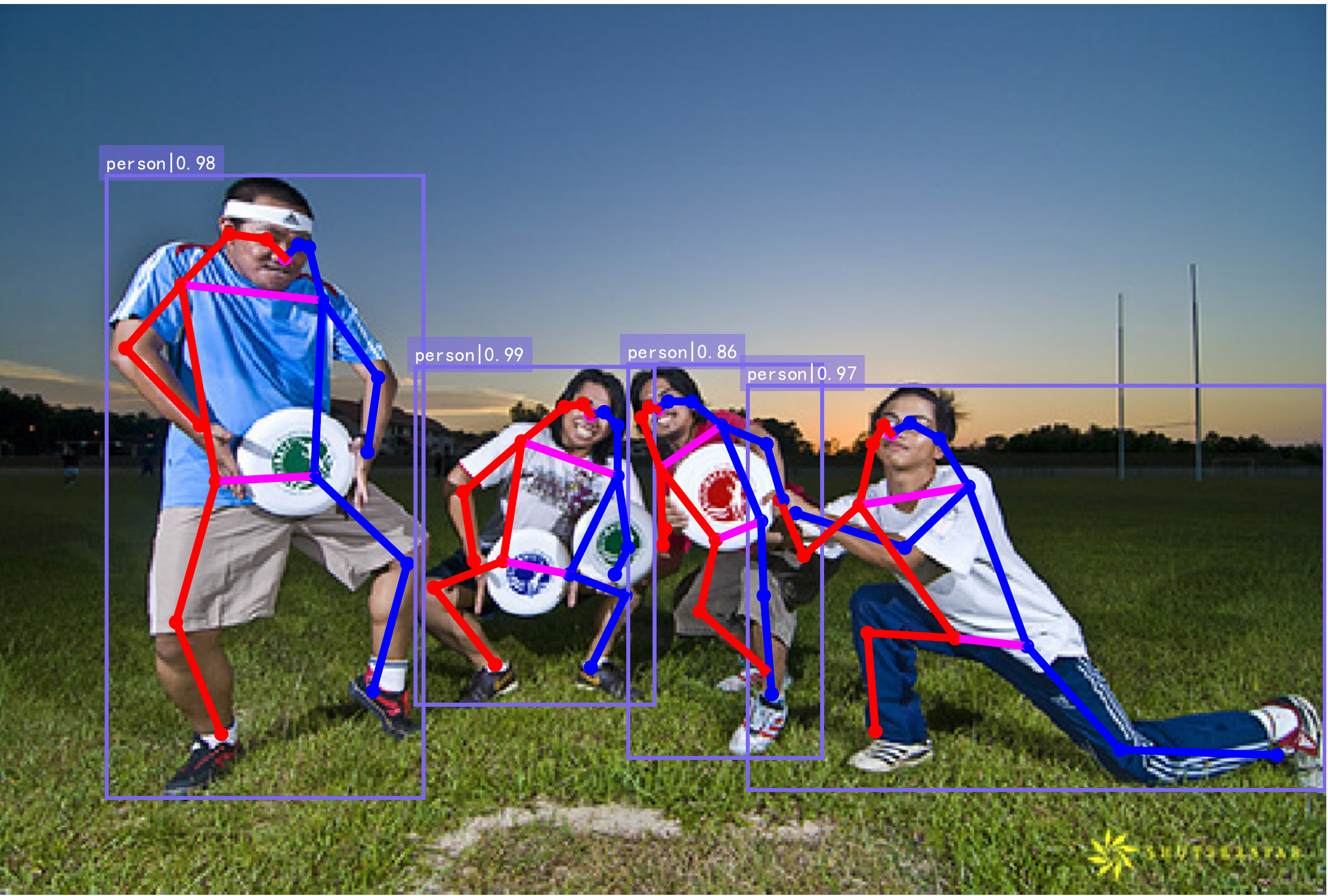}
    \label{fig8_6}
  }
  \caption{Some location-sensitive visual recognition results on the MS-COCO validation set. As discussed in Section~\ref{Segmentation}, the contour of the `motorcycles' in the figure cannot be well represented by the polar coordinate system in PolarMask.} 
  \label{fig:qualitative}
\end{figure*}


\subsection{Instance Segmentation}
\label{Segmentation}

\noindent \textbf{Comparisons to SOTA.} We show the instance segmentation inference results evaluated on the MS-COCO \textit{test-dev} set~\cite{lin2014microsoft} on Table~\ref{tab:segmentation}. LSNet achieves a mask AP of $38.0\%$ and $40.2\%$ using the single-scale and multi-scale testing protocols, respectively, surpassing all published contour-based methods to the best of our knowledge, and the accuracy is even competitive among the pixel-based approaches.

\noindent \textbf{Comparisons with PolarMask~\cite{xie2020polarmask}.} It is interesting to further compare our method with PolarMask, the previous best contour-based approach for instance segmentation. The major difference is that PolarMask assumed the entire object boundary to be seen by the anchor point, but this may not be the case especially for some complicated objects. Once the ray along some direction intersects with the border more than once, the method considered only one and thus incurred accuracy loss (a typical example is the ‘motorcycle’ contour in Figure~\ref{fig:qualitative}). In our approach, this issue is solved by ranking the landmarks more flexibly, being compatible to complicated shapes.

\noindent \textbf{The Number of Landmarks.} LSNet represents each instance using a polygon. Using a larger number of landmarks improves the upper-bound of accuracy, but can also incur heavy computational costs and cause the landmark prediction module difficult to be optimized. To choose a proper number of landmarks, we refer to the ground-truth masks of the MS-COCO validation set and quantize each mask into a polygon that best describes it. We find that using $18$, $36$, and $72$ landmarks achieves APs of $89.0\%$, $97.4\%$, and $99.2\%$, respectively, and we consider $N=36$ to be a nice tradeoff.

\subsection{Human Pose Estimation}
\label{section:pose}

\noindent \textbf{Comparisons to SOTA.} Unlike most of the human pose estimation methods that predict the keypoints using the heatmaps, LSNet predicts the keypoints using regression only. In the experiment, we use the object bounding boxes ('obj-box') and keypoint-boxes ('kps-box') to assign training samples, respectively. We will give a detailed discussion of the difference between the two methods in the Appendix~\ref{appendix_b}. On the MS-COCO \textit{test-dev} set, LSNet reports an AP of $55.7\%$ w/ obj-box and $59.0\%$ w/ kps-box, respectively, which outperform CenterNet-reg~\cite{zhou2019objects} with the Hourglass-104 backbone. However, LSNet does not perform as well as the heatmap-based methods, and we analyze the reason as follows.

\noindent \textbf{Error Analysis.} We can observe that the LSNet struggles particularly in the high OKS regimes, \textit{e.g.}, compared to Pose-AE~\cite{newell2017associative}, the deficit of AP$_{50}$ (for LSNet w/ obj-box) is $3.3\%$ while that of AP$_{75}$ grows to $9.0\%$. Note that using keypoint regression is not as accurate as using the heatmaps for refinement, and thus LSNet is less sensitive in the pixel-level prediction. However, the AP metric of pose estimation is largely impacted by this factor. To show this, we artificially add an average deviation of $1$, $2$, and $3$ pixels to the prediction results of CenterNet-jd~\cite{zhou2019objects} (with a backbone of Hourglass-104). The AP on the MS-COCO validation set is significantly reduced from $64.0\%$ (corresponding to the test AP of $63.0\%$ in Table~\ref{tab:pose}) to $61.1\%$, $53.4\%$, and $44.0\%$, respectively.

On the other hand, we use the heatmaps produced by CenterNet-jd (Hourglass-104) to refine the prediction of LSNet w/ obj-box. As a result, the AP on the MS-COCO validation set is improved from $56.5\%$ (corresponding to the test AP of $55.7\%$ in Table~\ref{tab:pose}) to $60.7\%$. This suggests that LSNet still needs further manipulation of high-resolution features towards higher pixel-level accuracy.

\noindent \textbf{The Benefit of LSNet.} Despite the relatively weak pixel-level localization, LSNet (w/ obj-box) enjoys the ability of perceiving multi-scale human instances, many of which are not annotated in the dataset. Some examples are shown in the right side of the Table~\ref{tab:pose}. Since the ground-truth is not available to evaluate the impact, we refer to the heatmaps of CenterNet-jd (Hourglass-104) to deliberately remove these `false positives'. Consequently, AP is further improved from $60.7\%$ to $63.0\%$, comparable with the heatmap-based methods, though the improvement seems less meaningful.

\subsection{Qualitative results for LSNet}
We show some visualized results of LSNet in Figure~\ref{fig:qualitative}, inlcuding object detection, instance segmentation and human pose estimation. Please refer to the appendix for more qualitative results.

\section{Conclusions}

This paper unifies three location-sensitive visual recognition tasks (object detection, instance segmentation, and human pose estimation) using the location-sensitive network (LSNet). The key module that supports the framework is a novel cross-IOU loss that is friendly to receiving supervision from multiple scales. Equipped with a pyramid DCN, LSNet achieves the state-of-the-art performance on anchor-free detection and segmentation. This work suggests that using keypoints to define and localize objects is a promising direction, and we hope to extend our approach to achieve a stronger ability of generalization.

\hspace{0.5ex}
\section*{APPENDIX}

\appendix
\renewcommand{\appendixname}{Appendix~\Alph{section}}

\section{The Softened Prediction Mechanism for Cross-IOU Loss}
\label{appendix_a}
\begin{figure}[h]
  \centering 
  \includegraphics[width=0.3\textwidth]{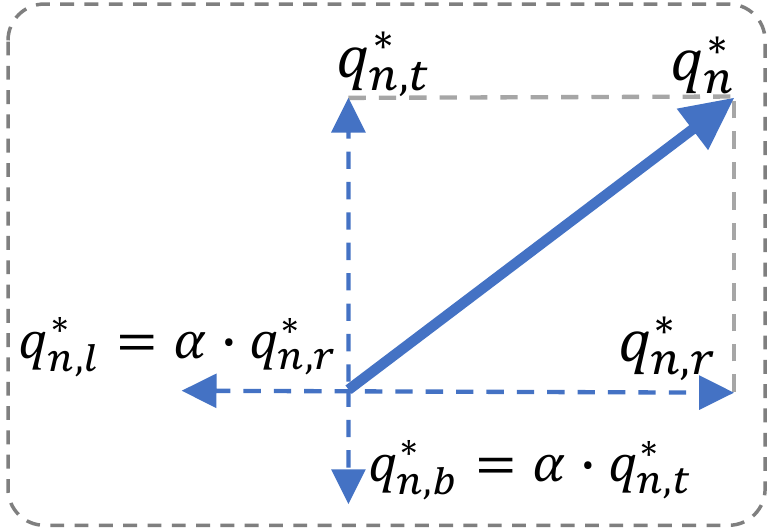}
  \caption{We use four components to represent a vector, and all the components are greater than $0$.}
  \label{fig:vector}
\end{figure}

As mentioned in Section~\ref{approach:localization}, the form of $\mathcal{L}_\mathrm{cIOU}$ (Equation 1) can cause the gradients over $p_{n,x}$ and $p_{n,y}$ to be $0$ when the corresponding dimensions of prediction and ground-truth are of different signs. To solve this problem, we predict four components for each offset vector, as shown in Figure~\ref{fig:vector}. Then $\mathbf{q}_n$ can be rewritten as: $\mathbf{q}_n=\left[q_{n,t},q_{n,l},q_{n,b},q_{n,r}\right]$, where $q_{n,l},q_{n,r},q_{n,t},q_{n,b}$ are all greater than $0$. On the other hand, when transforming $\mathbf{q}_n^\star$ into the cross-coordinate system, as shown in Figure~\ref{fig:vector}, we assign the minimum sides ($q_{n,l}^\star$ and $q_{n,b}^\star$) the non-zero value, which are $\alpha$ times the corresponding maximum sides ($q_{n,t}^\star$ and $q_{n,r}^\star$), where $0<\alpha<1$. In all our experiments, we set $\alpha=0.2$.

During inference, we transform the predicted offset vectors from the cross-coordinate system into the rectangular coordinate system by taking the maximum value in the horizontal and vertical direction, respectively, \textit{i.e.}, $\mathbf{q}_n=\left[\max\{q_{n,t},q_{n,b}\},\max\{q_{n,l},q_{n,r}\}\right]$.

\label{appendix_a}

\section{Assign Samples by Keypoint-boxes to Improve Human Pose Estimation}
\label{appendix_b}
In Section~\ref{section:pose}, we mainly discuss the characteristics of using the object bounding boxes ('obj-box') to assign training samples, which lets LSNet enjoy the ability of perceiving multi-scale human instances especially for small human instances, many of which are not annotated in the dataset. In this section, we mainly discuss the characteristics of using the keypoint-boxes ('kps-box', bounding box generated by the topmost, leftmost, bottommost and rightmost keypoints of an objects) to assign training samples. Compared with the former, the later will no longer treat the human instances that only have the object bounding box annotations but lack of pose annotations as positive samples. This makes the network pay more attention to learn the human instances that have the pose annotations, which helps to improve the AP score. As shown in Table~\ref{tab:pose}, LSNet using keypoint-boxes reports an AP of $59.0 \%$, an improvement of $3.3 \%$ over $55.7 \%$, achieved by LSNet using object bounding boxes.

\begin{figure*}[!t]
  \centering 
  \subfigure{ 
    \includegraphics[height=0.25\textwidth,width=0.15\textheight]{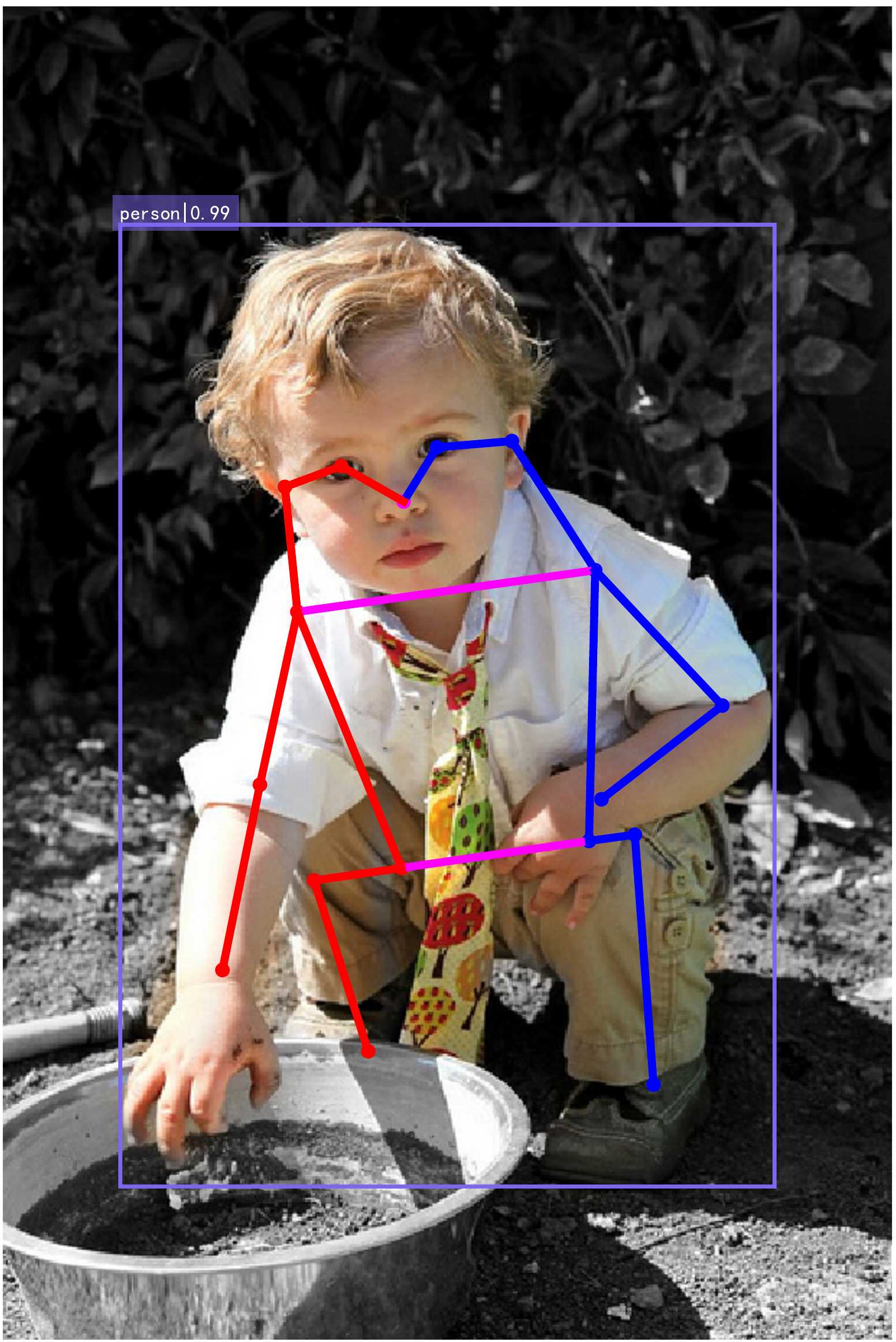}
  }
  \subfigure{ 
    \includegraphics[height=0.25\textwidth,width=0.15\textheight]{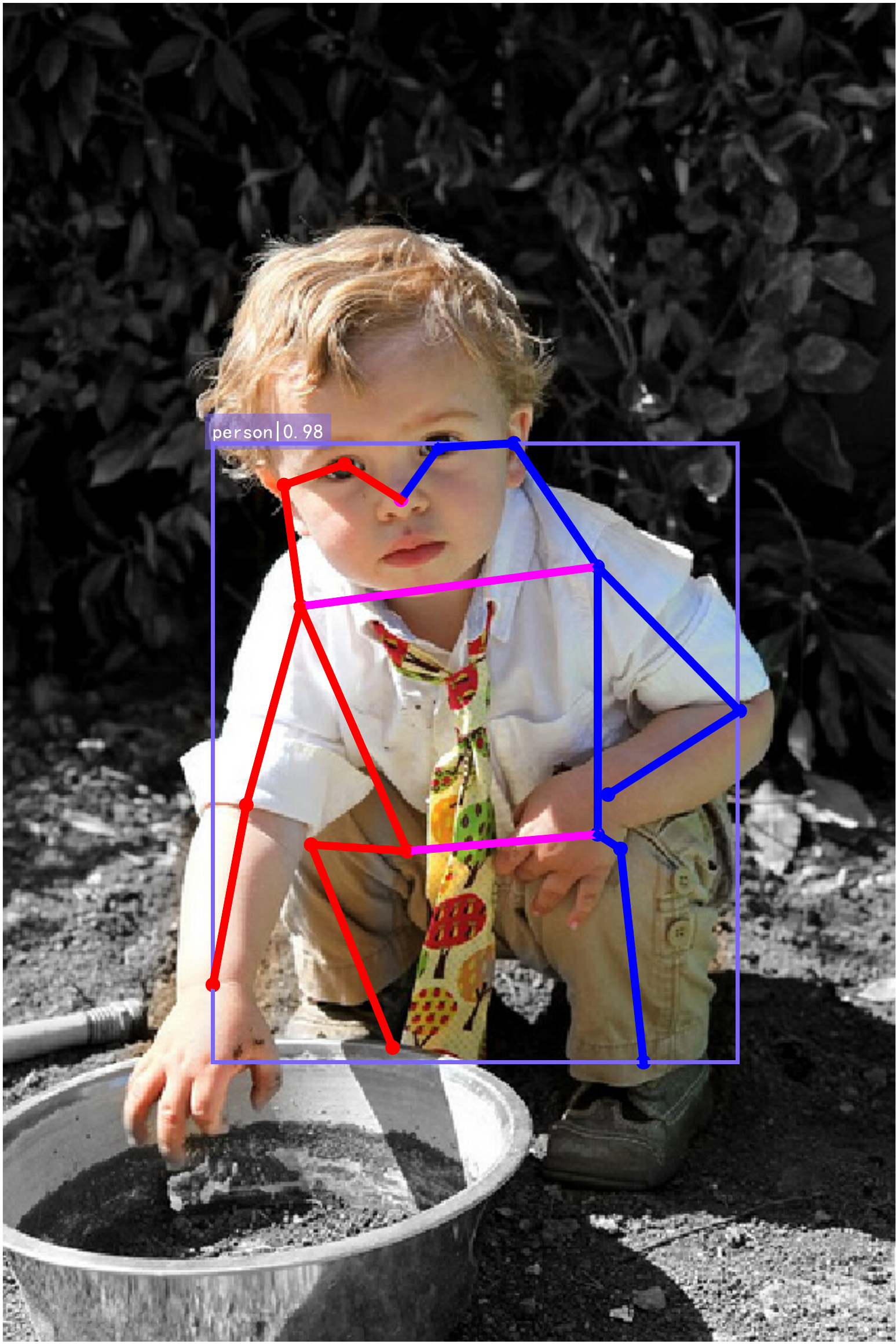}
  }
  \subfigure{ 
    \includegraphics[height=0.25\textwidth,width=0.15\textheight]{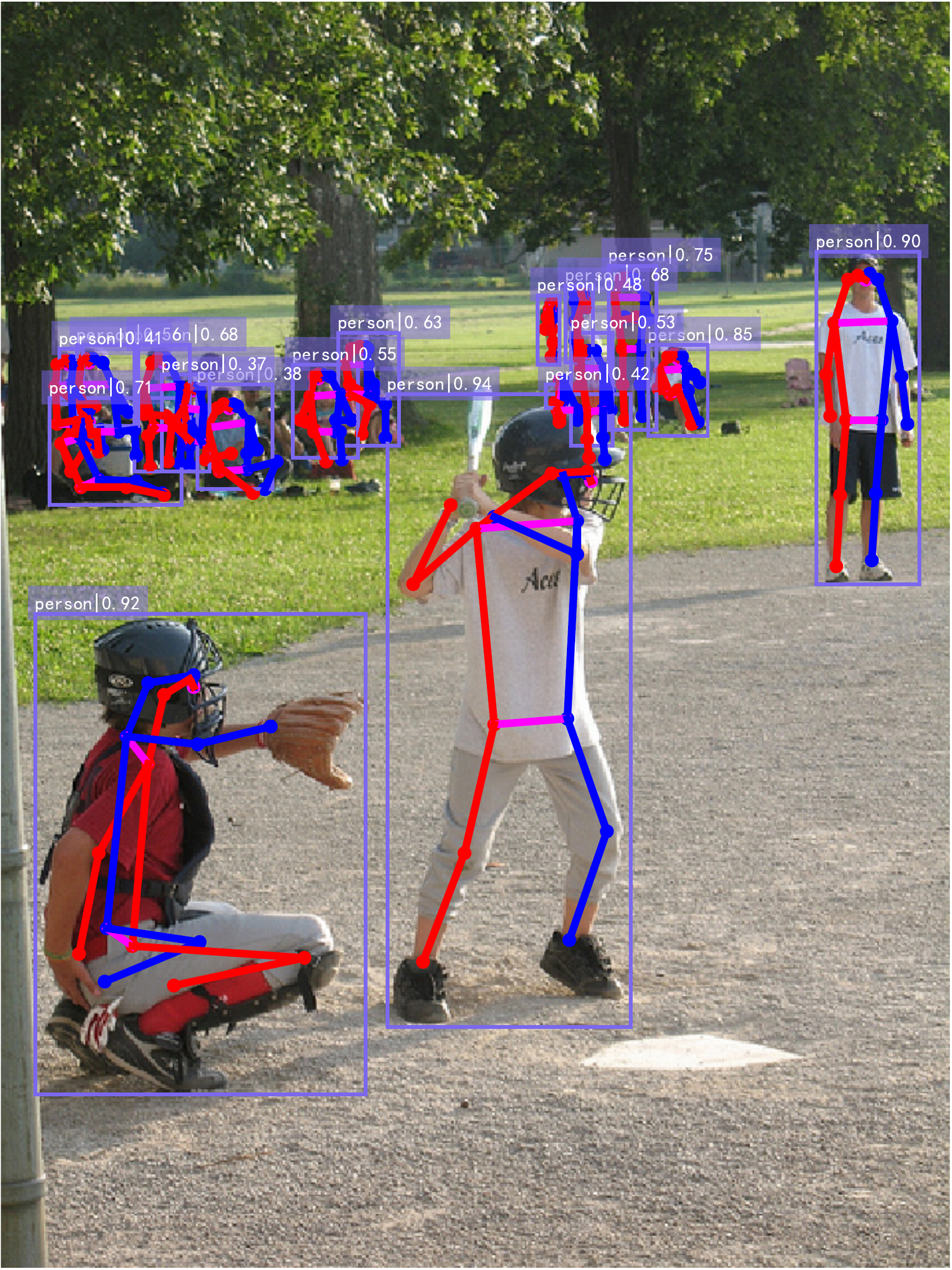}
  }
  \subfigure{ 
    \includegraphics[height=0.25\textwidth,width=0.15\textheight]{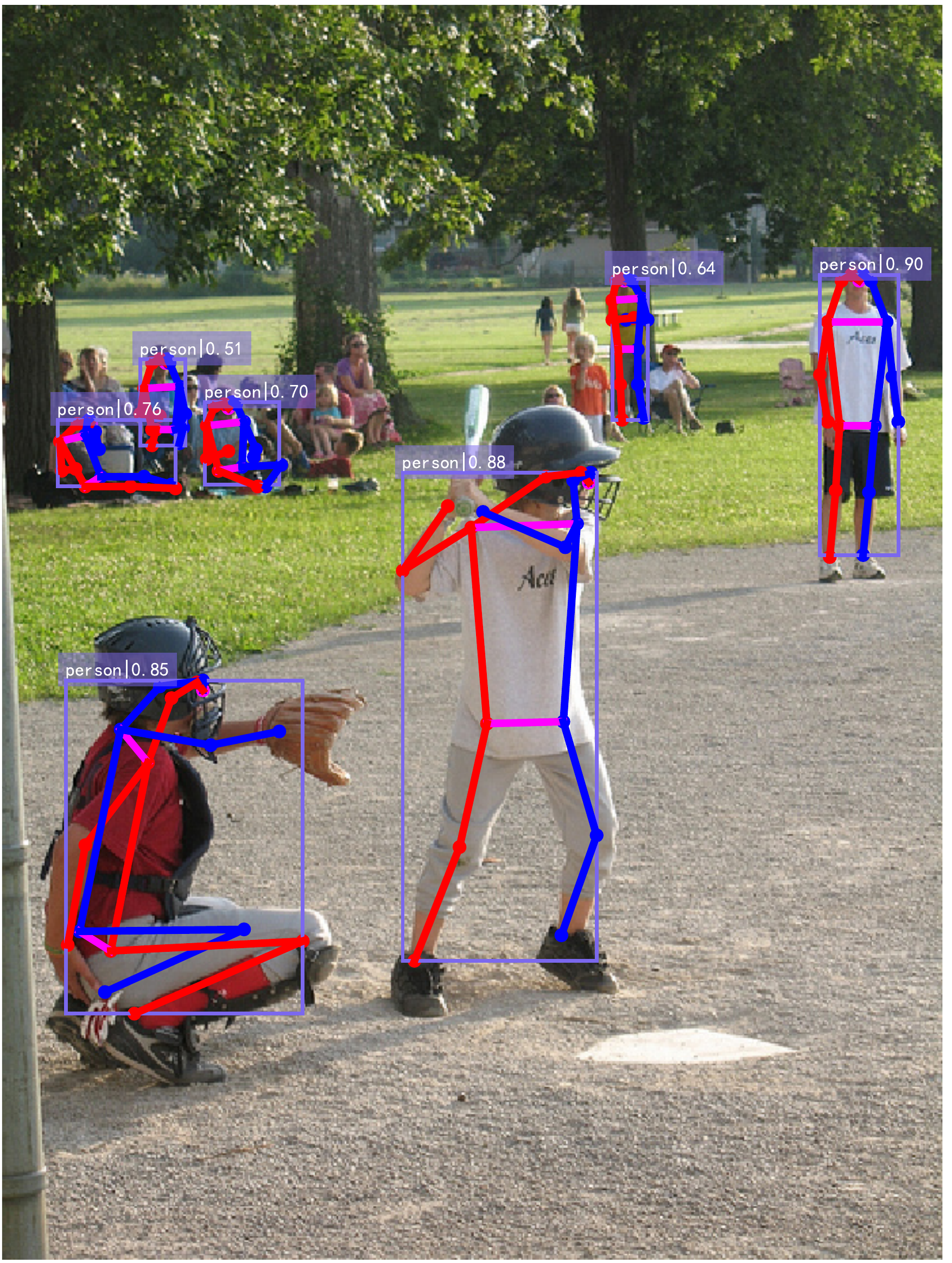}
  }
  \subfigure{ 
    \includegraphics[height=0.28\textwidth,width=0.312\textheight]{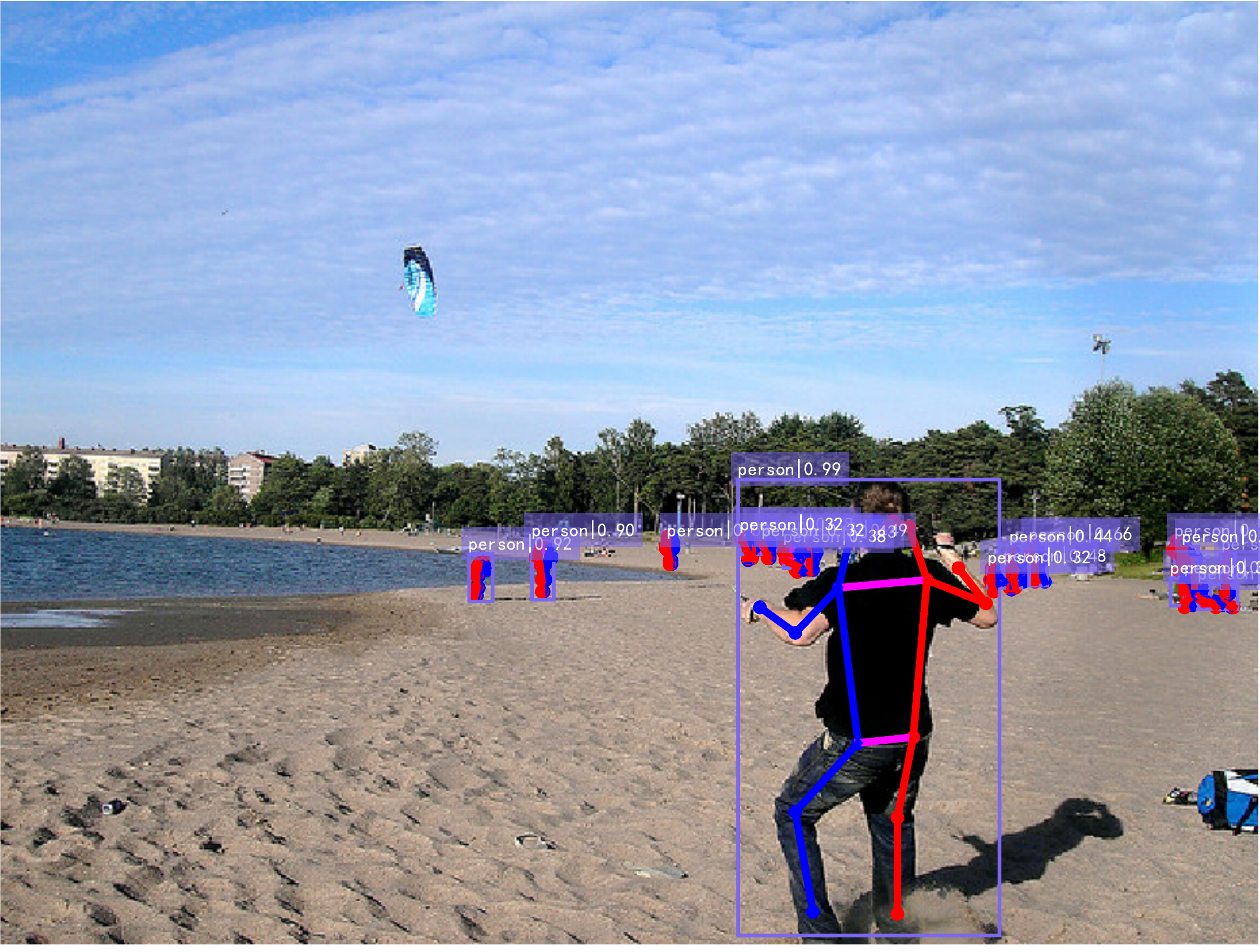}
  } 
  \subfigure{ 
    \includegraphics[height=0.28\textwidth,width=0.312\textheight]{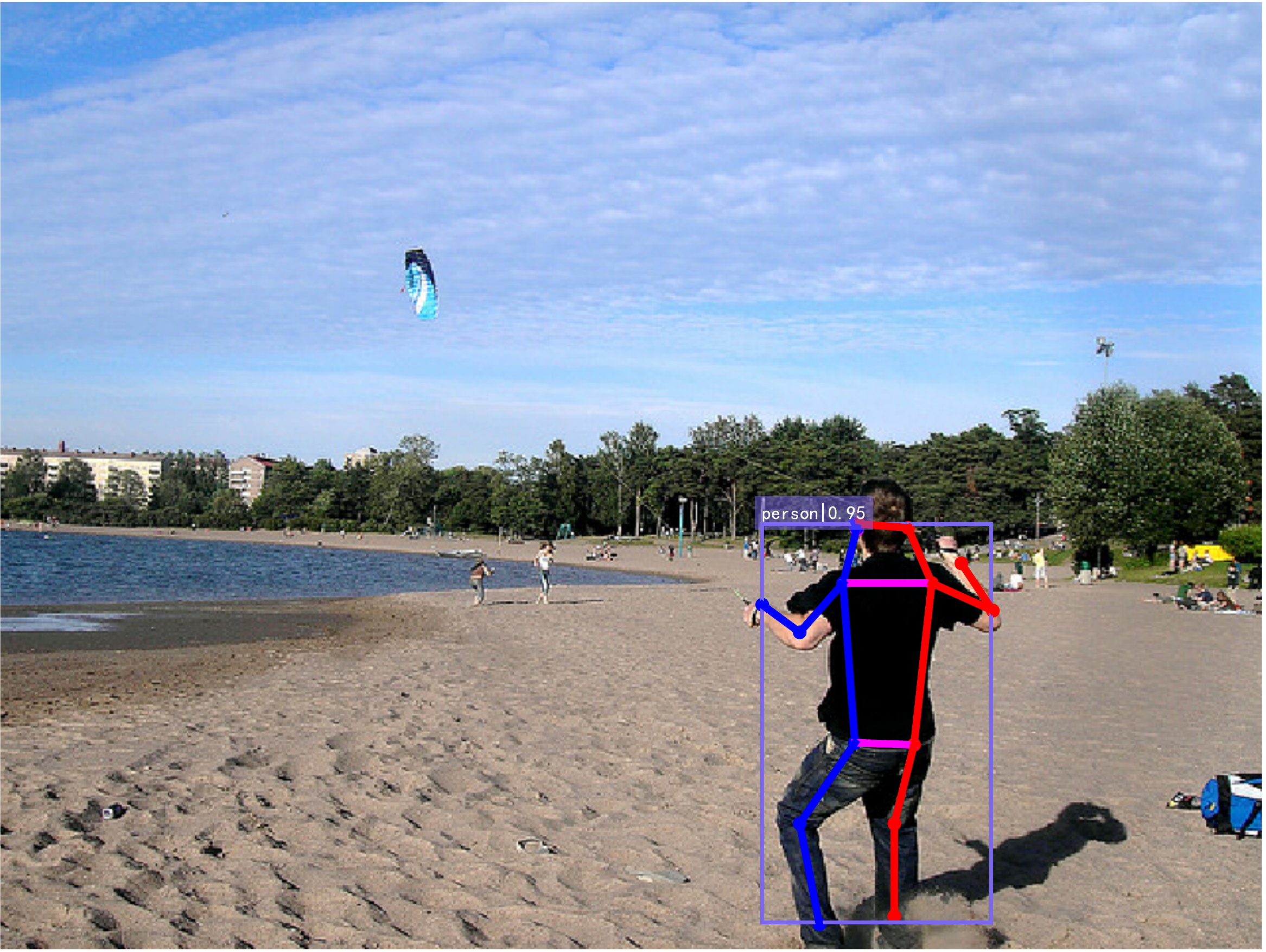}
  } 
  \subfigure{ 
    \includegraphics[height=0.28\textwidth,width=0.312\textheight]{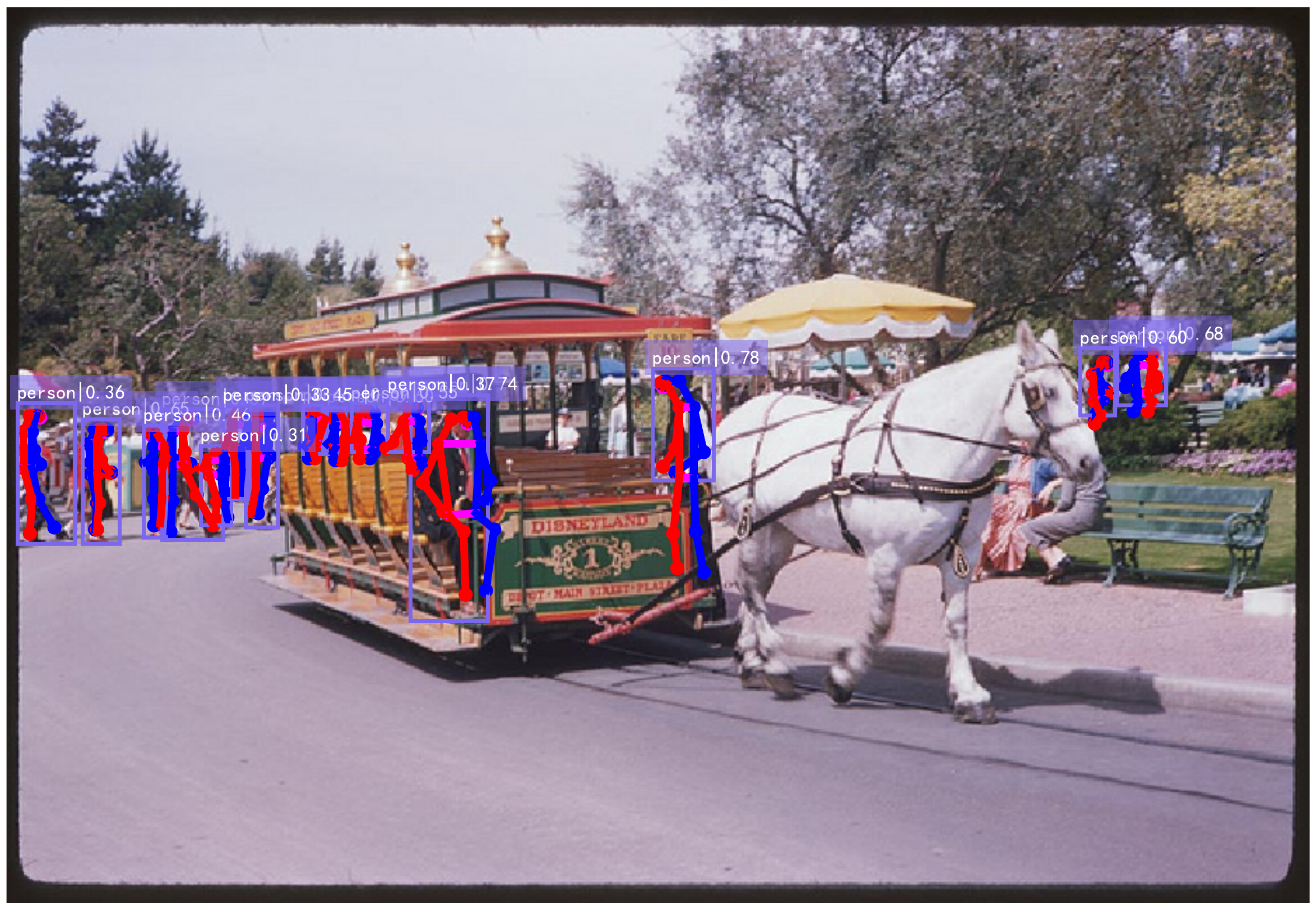}
  } 
  \subfigure{ 
    \includegraphics[height=0.28\textwidth,width=0.312\textheight]{figures/appendix/kps_kbox/000000016228.pdf}
  }
  \subfigure{ 
    \includegraphics[height=0.28\textwidth,width=0.31\textheight]{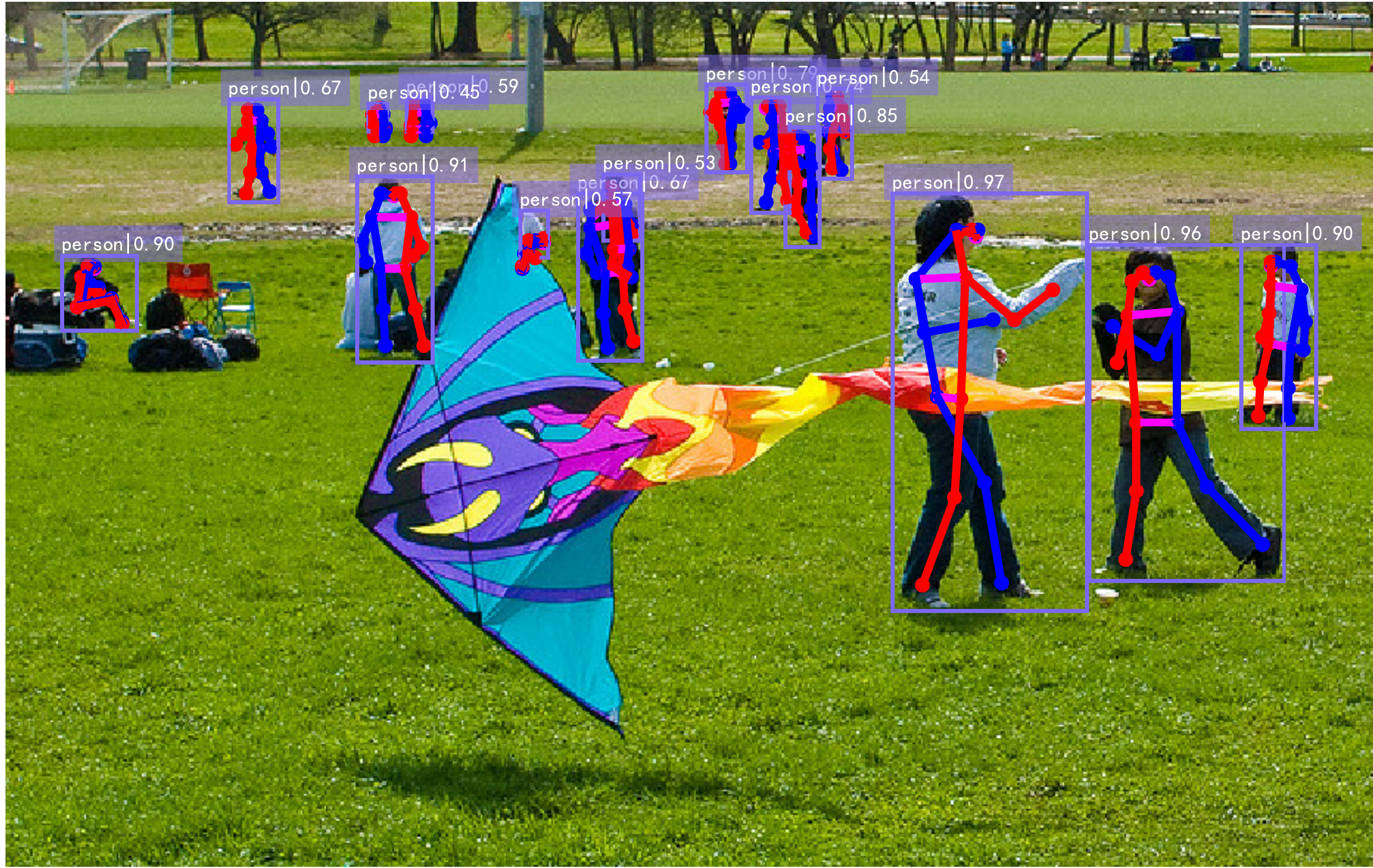}
  }
  \subfigure{ 
    \includegraphics[height=0.28\textwidth,width=0.312\textheight]{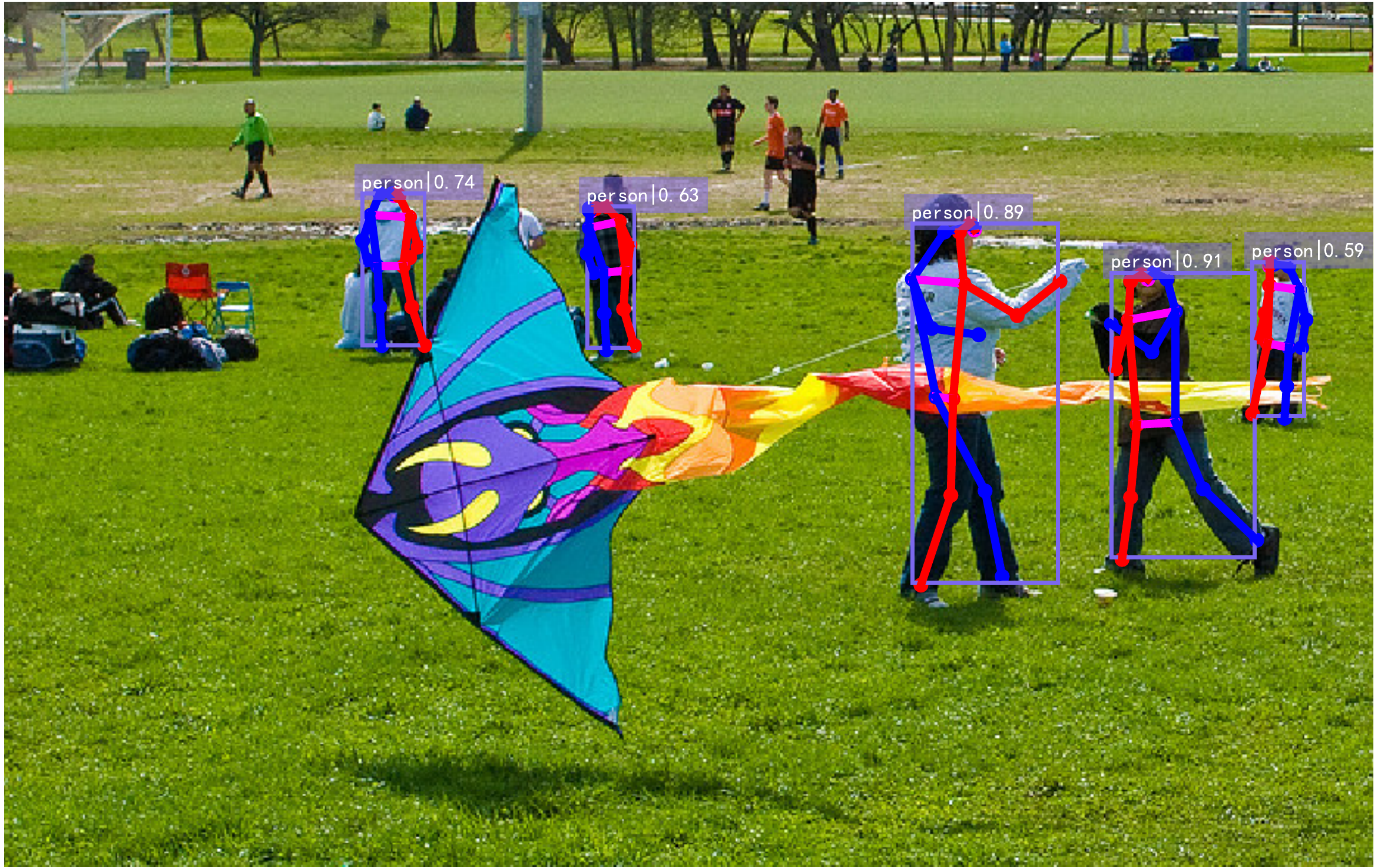}
  }
  \caption{Left: LSNet uses the object bounding boxes to assign training samples. Right: LSNet uses the keypoint-boxes to assign training samples. Although LSNet with keypoint-boxes enjoys a higher AP score, its ability of perceiving multi-scale human instances is weakened.} 
  \label{fig:pose_appe}
\end{figure*}

However, we find that, with the `improved' AP score, the ability of the algorithm at perceiving multi-scale human instances is weakened. As shown in Figure~\ref{fig:pose_appe}, the modified algorithm mostly fails to detect the small person instances. This proves that the annotations of the dataset is biased.

{\small
	\bibliographystyle{ieee}
	\bibliography{egbib}
}

\end{document}